\documentclass[sigconf,screen]{acmart}
\usepackage[nohyperlinks,nolist]{acronym}
\usepackage{textcomp}
\usepackage{multirow}
\usepackage{xcolor}
\definecolor{green}{RGB}{4,130,4}
\definecolor{violet}{RGB}{148, 1, 211}

\AtBeginDocument{%
  \providecommand\BibTeX{{%
    \normalfont B\kern-0.5em{\scshape i\kern-0.25em b}\kern-0.8em\TeX}}}

\setcopyright{acmcopyright}
\setcopyright{rightsretained}
\copyrightyear{2023}
\acmYear{2023}
\acmConference[ICAIL '23]{19th International Conference on Artificial
Intelligence and Law}{June 19--23, 2023}{Braga, Portugal}

\acmSubmissionID{141}

\begin{document}
%% The "title" command has an optional parameter,
%% allowing the author to define a "short title" to be used in page headers.
\title{MultiLegalSBD: A~Multilingual~Legal~Sentence~Boundary~Detection~Dataset}

%%
%% The "author" command and its associated commands are used to define
%% the authors and their affiliations.
%% Of note is the shared affiliation of the first two authors, and the
%% "authornote" and "authornotemark" commands
%% used to denote shared contribution to the research.
\author{Tobias Brugger}
\authornote{Both authors contributed equally to this research.}
\affiliation{
  \institution{University of Bern}
  \country{Switzerland}
}
\orcid{0009-0009-4617-1554}
\email{tobias.brugger@students.unibe.ch}

\author{Matthias Stürmer}
\affiliation{
  \institution{University of Bern}
  \country{Switzerland}
}
\affiliation{
  \institution{Bern University of Applied Sciences}
  \country{Switzerland}
}
\email{matthias.stuermer@unibe.ch}
\orcid{0000-0001-9038-4041}

\author{Joel Niklaus}
\authornotemark[1]
\affiliation{
  \institution{University of Bern}
  \country{Switzerland}
}
\affiliation{
  \institution{Bern University of Applied Sciences}
  \country{Switzerland}
}
\affiliation{
  \institution{Stanford University}
  \country{United States}
}
\email{joel.niklaus@unibe.ch}
\orcid{0000-0002-2779-1653}

%%
%% By default, the full list of authors will be used in the page
%% headers. Often, this list is too long, and will overlap
%% other information printed in the page headers. This command allows
%% the author to define a more concise list
%% of authors' names for this purpose.
\renewcommand{\shortauthors}{Brugger, Stürmer and Niklaus}

\begin{acronym}[UMLX]
    \acro{SBD}{Sentence Boundary Detection}
    \acro{CRF}{Conditional Random Fields}
    \acro{NN}{Neural Network}
    \acro{BiLSTM}{Bidirectional Long Short-Term Memory}
    \acro{LSTM}{Long Short-Term Memory}
    \acro{CNN}{Convolutional Neural Network}
    \acro{SJP}{Swiss-Judgement-Prediction}
    \acro{NLP}{Natural Language Processing}
    \acro{MLP}{Multi-Legal-Pile}
    \acro{NSR}{Negation Scope Resolution}
    \acro{R}{Recall}
    \acro{P}{Precision}
    \acro{F1}{F1-Score}
\end{acronym}

\begin{abstract}
  % Motivate why SBD is important
\ac{SBD} is one of the foundational building blocks of \ac{NLP}, with incorrectly split sentences heavily influencing the output quality of downstream tasks.
% SBD in legal text is difficult
It is a challenging task for algorithms, especially in the legal domain, considering the complex and different sentence structures used.
%The detection of sentence boundaries in legal text is a challenging task for algorithms, considering the complex and different sentence structures used compared to formal text. 
% we made a large dataset
In this work, we curated a diverse multilingual legal dataset consisting of over 130'000 annotated sentences in 6 languages.
% current systems are bad
Our experimental results indicate that the performance of existing SBD models is subpar on multilingual legal data.
% our models are better
We trained and tested monolingual and multilingual models based on CRF, BiLSTM-CRF, and transformers, demonstrating state-of-the art performance. 
% We outperform even in the zero shot setting
We also show that our multilingual models outperform all baselines in the zero-shot setting on a Portuguese test set. 
% We release our data, models and code
To encourage further research and development by the community, we have made our dataset, models, and code publicly available.
%We publicly release our dataset, models, and code for further use and research by the community.

\end{abstract}

%%
%% The code below is generated by the tool at http://dl.acm.org/ccs.cfm.
%% Please copy and paste the code instead of the example below.
%%


%%
%% Keywords. The author(s) should pick words that accurately describe
%% the work being presented. Separate the keywords with commas.

%% A "teaser" image appears between the author and affiliation
%% information and the body of the document, and typically spans the
%% page.
\begin{teaserfigure}
  \includegraphics[width=\textwidth,height=7.2cm]{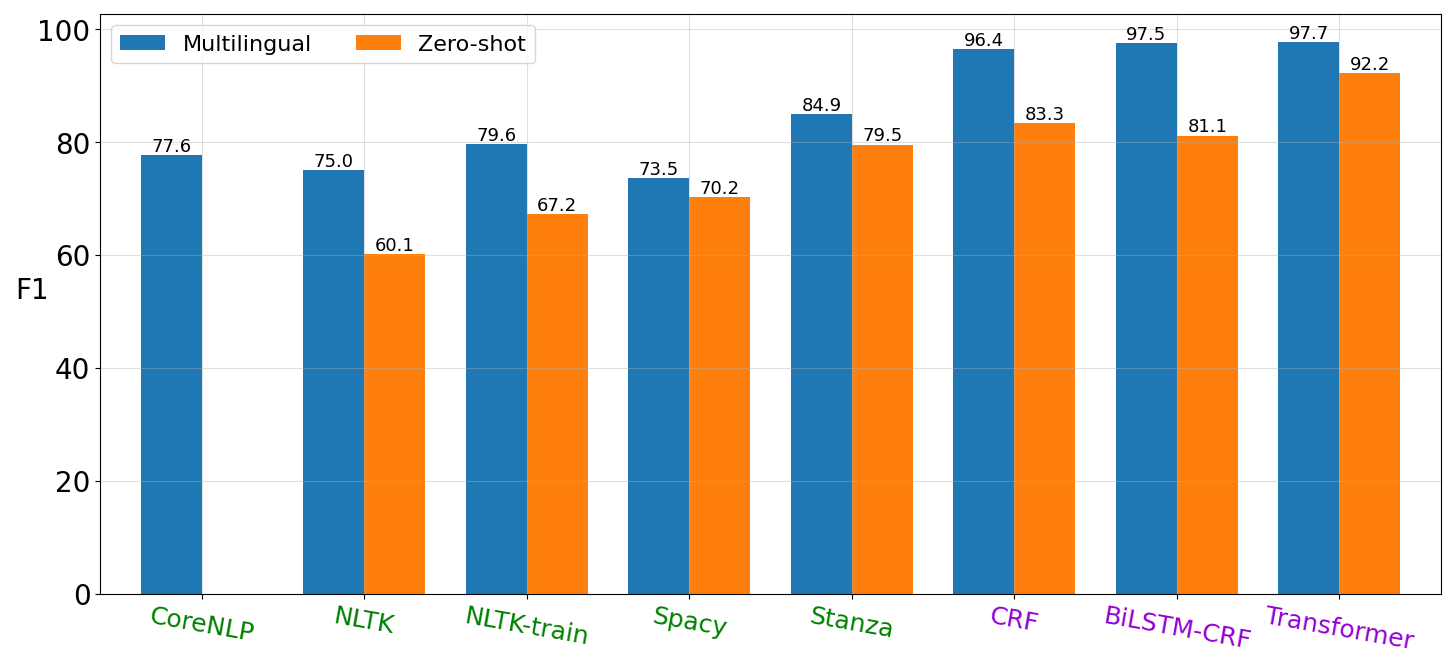}
  \caption{Mean performance comparison of baseline \textcolor{green}{(green)} and our multilingual models \textcolor{violet}{(violet)} on multilingual legal text (French, Italian, Spanish, English, and German) and a zero-shot experiment on Portuguese data}
  \Description[]{Values are provided in Table 3. The baseline models achieve F1-scores between 73.5 and 84.9 on multilingual texts and 60.1 to 79.5 on Portuguese texts. Stanza is the best baseline model overall. The multilingual models achieve F1-scores between 96.4 to 97.7 on the multilingual text and 81.1 to 92.2 on unseen Portuguese texts. The transformer model outperforms the CRF and BiLSTM-CRF models on both tasks.}
  \label{fig:teaser}
\end{teaserfigure}

\received{16 February 2023}
\received[accepted]{12 April 2023}
\received[revised]{2 May  2023}

%%
%% This command processes the author and affiliation and title
%% information and builds the first part of the formatted document.
\maketitle

\acresetall

\section{Introduction}
\label{ch:Introduction}

% A lot of Growth in NLP
Recent methodological advances, e.g., transformers \cite{Vaswani.etal2017}, have lead to substantial progress in quality and performance of language models as well as growth in the general field of \ac{NLP}. This trend is also evident in legal NLP, with research papers increasing drastically in recent years \cite{Katz.etal2023}.

% Research in SBD is seen as solved
% SBD difficult in domain specific applications
Not as much attention and resources have been directed to the \ac{SBD} task, being viewed as solved by some, as high baseline performances can be achieved by utilizing simple lookup methods capturing frequent sentence-terminating characters such as periods, exclamations marks and question marks combined with hand-crafted rules \cite{Read}. This approach is feasible when applied to well-formed and curated text such as news articles. Noisier domain-specific data containing differently structured text combined with the ambiguity of many sentence-terminating characters \citep{Kiss, Gillick2009} -- e.g., the period occurring in abbreviations, ellipses, initials etc. as a non-terminating character -- often overwhelm the aforementioned methods and also more complicated off-the-shelf \ac{SBD} systems. This has been illustrated in a number of specific \ac{SBD} applications such as user-generated content \citep{Read, Gimpel.etal2011} as well as in the clinical \citep{Newman-Griffis.etal2016} and financial domain \citep{Mathew.Guggilla2019, Du}.

In legal documents, the aforementioned difficulties are increased with legal text consisting of smaller parts such as paragraphs, clauses etc., making it quite different from standard text. Furthermore, sentences are long and may contain complex structures such as citations, parentheses, and lists. These structures are often utilized to convey additional information to the reader (e.g., citations referencing another text) or formatting the text in a specific way (e.g., lists emphasizing ideas or increasing the readability of long paragraphs). However, these structures or special sentences do not follow a standard sentence structure, thus posing an additional challenge to \ac{SBD} systems, illustrated in several works on English \citep{Savelka, Sanchez} and German \citep{Glaser} legal documents.

% What is SBD
% legal text different from normal text because -> examples
% legal text more difficult because 
%\ac{SBD} has the goal of splitting a body of text into complete and useful sentences. It is a non-trivial task, especially in the legal domain. Legal text consists of long paragraphs, clauses etc., making it quite different from regular text. Furthermore, complex text blocks such as headings, citations, parentheses, and lists are often utilized to convey additional information to the reader (e.g., citations referencing another text) or formatting the text in a specific way (e.g., lists emphasizing ideas or increasing the readability of long paragraphs). However, these blocks do not follow a standard sentence structure, thus posing an additional challenge to \ac{SBD} systems.

\subsection{Motivation}
% SBD is important to mitigate errors flowing into next NLP tasks
%\ac{SBD} is often the initial pre-processing step required for further \ac{NLP} analysis of text. Therefore, it is vital to have a solid \ac{SBD} system in place to dampen the propagation of errors into higher-level text processing tasks, lowering the overall performance.
% SBD application examples
% machine translation EUROPARL corpus
% further examples: text summarization, information extraction
%An important example of such a task is found in the curation of the EUROPARL corpus \cite{koehnEuroparlParallelCorpus2005}, a collection of the proceedings of the European Parliament in over 20 languages, for research in statistical machine translation. To create language translation pairs, \citet{koehnEuroparlParallelCorpus2005} had to align the same sentences in both languages, making it necessary to split the text into sentences beforehand. In their work, they mention the difficulty of \ac{SBD}, requiring specialized tools for each language, which were not readily available for all languages.
%Suboptimal \ac{SBD} most likely weakens the performance of the sentence alignment algorithm and therefore the overall quality of the corpus. A high-quality SBD system, especially one adapted to the legal domain and language, could improve the quality by a large margin. Other examples of \ac{NLP} tasks where \ac{SBD} plays a critical role, can be found in text summarization, Part-of-Speech-Tagging and Named Entity Recognition, all very pertinent to the legal domain. 

Having a reliable \ac{SBD} system is crucial for accurate \ac{NLP} analysis of text. Poor \ac{SBD} can result in errors propagating into higher-level text processing tasks, which hinders overall performance. For instance, the curation of the multilingual EUROPARL corpus required proper \ac{SBD} to align sentences in both languages for statistical machine translation. \citet{koehnEuroparlParallelCorpus2005} noted the difficulty of \ac{SBD} as it requires specialized tools for each language, which are not readily available for all languages. Inadequate \ac{SBD} weakens the performance of sentence alignment algorithms and reduces the quality of the corpus. Therefore, a high-quality \ac{SBD} system, especially one customized for the legal domain, can significantly improve performance.

Another example is \ac{NSR}, focusing on finding negation words (e.g., "not") in sentences and their impact on surrounding words' meaning. Negations are vital in text's semantic representation, reversing proposition values.
This is particularly useful in the legal domain, enabling models extracting information from documents to better understand input text meaning, such as recognizing court decisions' outcomes based on exact wording. \ac{NSR} models often require data split into sentences for labeling training data and application input, making a reliable SBD system crucial. Incorrect sentence predictions by the SBD system may significantly lower input data quality and model performance.
Proper \ac{SBD} is also crucial in other \ac{NLP} tasks such as Text Summarization, Part-of-Speech-Tagging, and Named Entity Recognition, all relevant in the legal domain.

%Another example, building on the former task, is Legal Judgement Prediction, where an AI model is trained to predict the verdict of a court decision given its facts. Such models have several applications, e.g., help lawyers by identifying their strengths and weaknesses or help judges and clerks to review or prioritize cases, therefore speeding up and improving the quality of judicial processes \cite{NiklausSwiss}.

%SBD bad in domain-specific tasks ->  Citations
%No research in different languages / multilingual
%However, current \ac{SBD} systems exhibit sub-optimal performance working on informal language and special domains such as user-generated content \citep{Read}. Further research \citep{Savelka, Glaser} emphasizes this problem on English and German legal text.
%To the best of our knowledge, there has not been any research on how to improve this issue in other languages such as French, Italian and Spanish and no multilingual approach has been investigated in the legal domain.

%write intro to RQ?

\subsection{Main Research Questions}
\label{sec:research_questions}
In this work, we pose and examine three main research questions:
\begin{description}
    \item[RQ1:] What is the performance of existing SBD systems on legal data in French, Spanish, Italian, English, and German?
    \item[RQ2:] To what extent can we improve upon this performance by training mono- and multilingual models based on CRF, BiLSTM-CRF, and transformers?
    \item[RQ3:] What is the performance of the multilingual models on unseen Portuguese legal text, i.e., a zero-shot experiment?
\end{description}

\subsection{Contributions}
The contributions of this paper are twofold:

\begin{enumerate}
    \item We curate and publicly release a large, diverse, high-quality, multilingual legal dataset (see Section \ref{ch:Dataset}) containing over 130'000 annotated sentence spans for further research in the community.
    \item Using this dataset, we showcase that existing SBD systems exhibit suboptimal performance on legal text in French, Italian, Spanish, English, and German. 
    We train and evaluate state-of-the-art monolingual SBD models based on \ac{CRF}, BiLSTM-CRF and transformers, achieving F1-scores up to 99.6\%.
    We showcase the performance and feasibility of multilingual SBD models, i.e., trained on all languages, achieving F1-scores in the higher nineties, comparable or better than our monolingual models on each aforementioned language.
    In a zero-shot experiment, we demonstrate that it is possible to achieve good cross-lingual transfer by testing the multilingual models on unseen Portuguese legal text.
    We publicly release the datasets\footnote{\url{https://huggingface.co/datasets/rcds/MultiLegalSBD}}, all of our monolingual and multilingual models\footnote{\url{https://huggingface.co/models?search=rcds/distilbert-sbd} and \url{https://github.com/tobiasbrugger/MultiLegalSBD/tree/master/models}} (see Section \ref{ch:Results}) as well as our code\footnote{\url{https://github.com/tobiasbrugger/MultiLegalSBD}} for further use in the community.
\end{enumerate}

\section{Related Work}
\label{ch:RelatedWork}

In this section, we discuss the literature at our disposal. First, we look at works showcasing the need for more research in regard to SBD. Second, we take a look at works tackling the problem of \ac{SBD} in legal text in several languages. Lastly, we investigate SBD research in other domains and present multilingual datasets in the legal domain for thoroughness.

\citet{Read} questioned the status quo of SBD being "solved", especially in more informal language and special domains, by reviewing the current state-of-the-art \ac{SBD} systems on English news articles and user-generated content. 
The systems were able to reach F1-scores in the higher nineties for the former, however the performance on user-generated content weakened perceptibly with scores down to the lower nineties, showcasing the need for "a renewed research interest in this foundational first step in \ac{NLP}." \citep{Read}

\subsection{SBD in the Legal Domain}

\citet{Savelka} continued this research in the English language by curating a legal dataset, consisting of adjudicatory decisions from the United States. When testing existing systems on the dataset, they report F1-scores between 75\% and 78\%. Training or adapting these systems to the dataset improved their F1 score to the mid-eighties, which is still lower than their respective performance in more standard domains \citep{Read}, showcasing the subpar performance of state-of-the-art SBD in the English legal domain. To improve this issue, they trained a number of \ac{CRF} models as well as a model based on hand-crafted rules, reporting F1-scores of 79\% for the hand-crafted model and up to 96\% for the \acp{CRF}. Additionally, they developed a publicly available, comprehensive set of annotation-guidelines for sentence boundaries in legal texts which we used as a foundation for our guidelines.

\citet{Sanchez} experimented on the same dataset reporting an F1-score of 74\% using the Punkt Model \cite{Kiss}; adapting it to the dataset slightly improved performance. They also trained and evaluated \ac{CRF} and \ac{NN} models, reporting F1-scores up to 98.5\% and 98.4\% respectively. Our multilingual models achieve F1-scores between 95.1\% and 97\% on the same dataset.

Similarly, \citet{Glaser} curated a German legal dataset, split into laws and judgements; a similar distribution is used in our work.
They established a baseline performance of existing \ac{SBD} systems and compared it to \ac{CRF} and \ac{NN} models trained on the aforementioned dataset. Their findings outline F1-scores between 70\% to 78\% for off-the-shelf systems, supporting the view that the performance of existing SBD system is subpar on legal data. The \acp{CRF} and \acp{NN} models achieve F1-scores up to 98.5\%. However, a significant decrease in performance was reported, when applying them to previously unseen German legal texts with scores down to 81.1\%. Our multilingual models showcase F1-scores between 91.6\% to 97.6\% on the German dataset.

\subsection{SBD in Other Domains}
In the financial domain, \citet{Du} experimented with \ac{BiLSTM} models combined with a \ac{CRF} layer as well as the transformer-based model BERT \citep{Devlin.etal2019} and compared their performance, approaching SBD as a sequence labelling task to extract useful sentences from noisy financial texts. They demonstrate that BERT significantly outperforms BiLSTM-CRFs across all evaluation metrics, including F1-scores. In their work they also underline the fact that "SBD has received much less attention in the last few decades than some of the more popular subtasks and topics in NLP."

\citet{SchweterAhmed} compared the performance of \acp{LSTM}, \acp{BiLSTM} and \acp{CNN} to OpenNLP \footnote{https://opennlp.apache.org/} in an \ac{SBD} task on the Europarl \citep{koehnEuroparlParallelCorpus2005}, SETimes \citep{Tiedemann2012} and Leipzig Corpora \citep{Goldhahn.etal2012} containing around 10 different languages, showcasing the use of their models as robust, language-independent SBD systems.

\subsection{Multilingual Datasets in the Legal Domain}
\citet{niklaus_lextreme_2023} present LEXTREME, a novel multilingual benchmark dataset containing 11 datasets in 24 languages, designed to evaluate natural language processing models on legal tasks. The authors assess five prevalent multilingual language models, providing a benchmark for researchers to use as a basis for comparison.
\citet{Savelka.etal2021} investigate the application of multilingual sentence embeddings in sequence labeling models to facilitate transfer across languages, jurisdictions, and other legal domains. They demonstrate encouraging outcomes in allowing the reuse of annotated data across various contexts, which leads to the development of more resilient and generalizable models. Additionally, they create a vast dataset of newly annotated legal texts using these models.
\citet{chalkidis_multieurlex_2021} introduce MultiEURLEX, a multilingual and multilabel legal document classification dataset containing 65000 EU Laws.
\citet{aumiller_eur-lex-sum_2022} present a EurLexSum, a multilingual summarization dataset curated from Eur-Lex data.
\citet{niklaus_swiss-judgment-prediction_2021, niklaus_empirical_2022} introduce Swiss-Judgment-Prediction, a multilingual judgment prediction dataset from the Federal Supreme Court of Switzerland. 

\section{Dataset}
\label{ch:Dataset}

%In an effort to create a diverse set of multilingual legal data and obtain an in-depth look at the performance of legal \ac{SBD} in a multitude of languages, we have annotated sentence spans for three distinct datasets, one each for French, Italian, and Spanish.
%Each dataset contains around 20,000 sentences, split evenly between judgments and laws, chosen to capture a diverse selection of different areas of law. The laws contain the Constitution, part of the Civil Code as well as part of the Criminal Code. Since the Constitution is a relatively small subset, it was used for evaluation only. The judgments contain court decisions from various sources and legal areas. Furthermore, we have annotated a smaller Portuguese dataset containing around 1800 sentences, with the same relative subsets as described above. This dataset was used for zero-shot experiments.

We annotated sentence spans for three diverse multilingual legal datasets in French, Italian, and Spanish, each containing approximately 20,000 sentences evenly split between judgments and laws. We chose a variety of legal areas to capture a broad selection. The laws included the Constitution, part of the Civil Code, and part of the Criminal Code, with the Constitution used only for evaluation. The judgments comprised court decisions from various legal areas and sources. We also annotated a smaller Portuguese dataset with approximately 1800 sentences, divided into the same subsets as the other datasets. This dataset was used for zero-shot experiments.

Additionally, we standardized and integrated two publicly available datasets, an English collection of legal texts \cite{Savelka}, consisting of Adjudicatory Decision from the United States as well as a German dataset \cite{Glaser}, comprising laws and judgments, into our dataset to further increase its diversity.

%The sentence length distribution of our dataset is illustrated in Fig. \ref{fig:senDist}, showcasing the relative frequency of sentence length in tokens for laws and judgments, with a bin size of 5. 
%The sentences were tokenized using an aggressive tokenizer, described further below, resulting in a larger amount of tokens per sentences than usual.\footnote{Single-token sequences like "Facts" result in one-token length sentences.} We do not show sentences longer than 101 tokens for clarity, with only \textasciitilde{}2\% (2634) sentences being longer. Only 26 sentences are longer than 512 tokens.

\begin{figure}
    \includegraphics[width=\columnwidth]{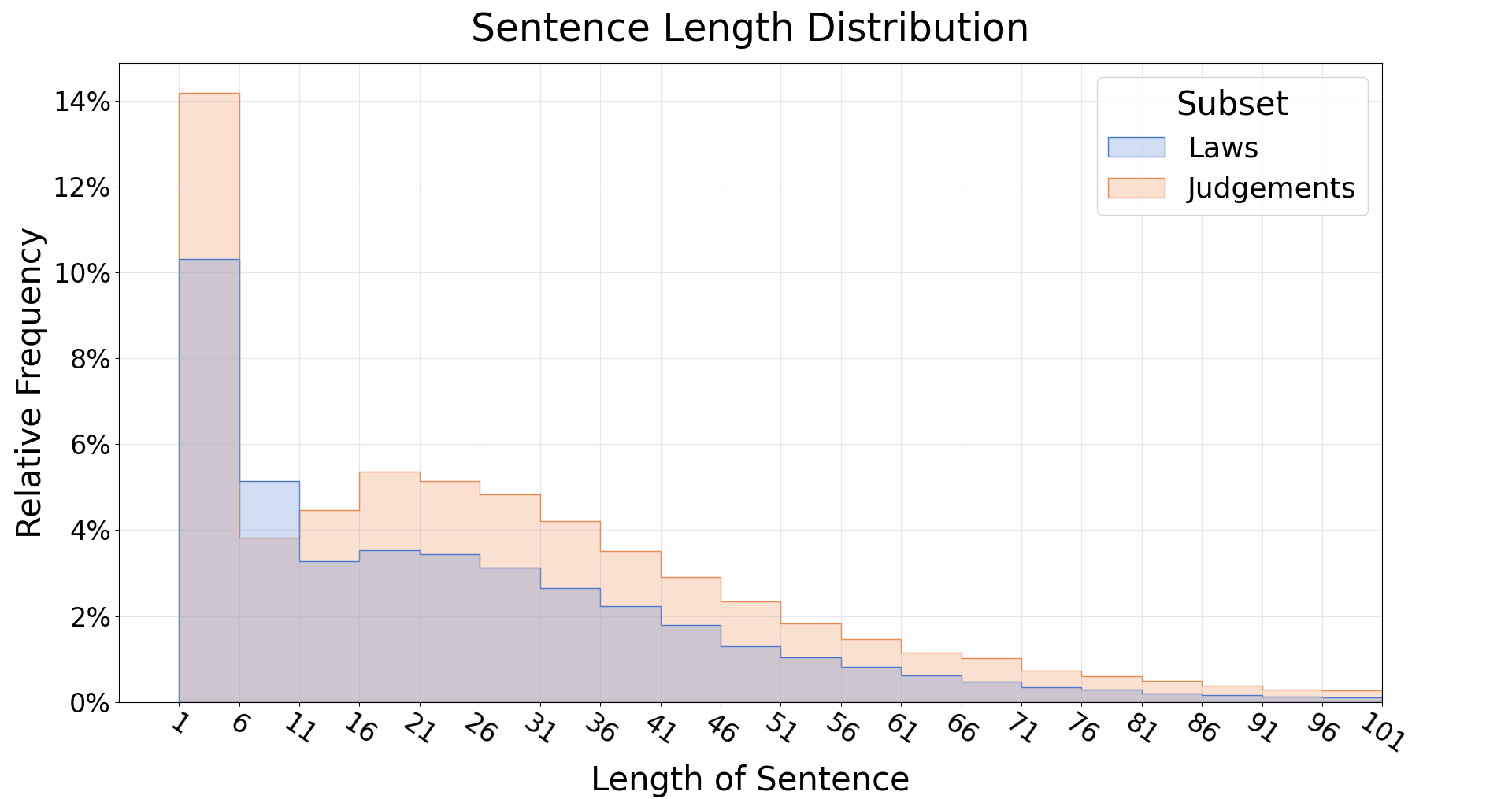}
    \caption{Sentence length distribution in tokens}
    \Description[]{Figure 2 shows the sentence length distribution in tokens of the law and judgments subsets of our dataset as a histogram. Both subsets have a relatively high amount of sentences between 1-5 tokens. The laws contain more sentences between 6-11 tokens, however the Judgments contain more and longer sentences from 11 tokens onwards compared to the laws. For both subsets, the amount of sentences slowly drops off after 21 tokens.}
    \label{fig:senDist}
\end{figure}

Figure \ref{fig:senDist} illustrates the sentence length distribution of our dataset, showing the relative frequency of sentence length in tokens for laws and judgments, with a bin size of 5. We used an aggressive tokenizer, resulting in a larger number of tokens per sentence than usual. For clarity, we did not include sentences longer than 101 tokens, which comprised only \textasciitilde{}2\% (2634) of the sentences. Only 26 sentences were longer than 512 tokens.

% The longest sentence in our dataset is 2412 tokens.

For each language, we used random sampling to split the dataset into three parts: train, test and validation. The test and validation splits each contain 20\% of the dataset. Every model is trained on the train split, and we report their performance on the test split.
Selected statistics and information about the dataset are in Table \ref{tab:DatasetStats}.
% and links to our data sources are listed in Appendix \ref{ss:datasetSources}. 

%\begin{description}
%    \item \href{https://wipolex.wipo.int/en/main/home}{Wipolex}
%    \item \href{http://www.jus.unitn.it/cardozo/Obiter_Dictum/}{Jus.unitn.it}
%    \item \href{https://huggingface.co/datasets/swiss_judgment_prediction}{\ac{SJP}}
%    \item \href{https://huggingface.co/datasets/joelito/Multi_Legal_Pile}{\ac{MLP}}
%    \item \href{https://github.com/sebischair/Legal-Sentence-Boundary-Detection}{\citet{Glaser}}
%    \item \href{https://github.com/jsavelka/sbd_adjudicatory_dec}{\citet{Savelka}}
%\end{description}

\begin{table*}
\centering
\caption{Statistics on datasets per language and subset}
\label{tab:DatasetStats}
\begin{tabular}{llrrrl} 
\toprule
Language & Subset & Sentences & Tokens & \# of Documents & Source \\ 
\midrule
French & Judgments & 9971 & 342469 & 315 & \href{https://huggingface.co/datasets/swiss_judgment_prediction}{\citet{niklaus_swiss-judgment-prediction_2021}} \\
 & Laws & 11055 & 334453 & 3 & \href{https://wipolex.wipo.int/en/main/home}{Wipolex}\\ 
\hline
Italian & Judgments & 10129 & 340041 & 183 + 60 & \href{https://huggingface.co/datasets/swiss_judgment_prediction}{\citet{niklaus_swiss-judgment-prediction_2021}}  + \href{https://huggingface.co/datasets/joelito/Multi_Legal_Pile}{\ac{MLP}} \\
 & Laws & 10849 & 301466 & 3 &  \href{http://www.jus.unitn.it/cardozo/Obiter_Dictum/}{Jus.unitn.it}\\ 
\hline
Spanish & Judgments & 10656 & 356681 & 20 + 84 & \href{https://wipolex.wipo.int/en/main/home}{Wipolex} + \href{https://huggingface.co/datasets/joelito/Multi_Legal_Pile}{\ac{MLP}} \\
 & Laws & 11501 & 229240 & 3 & \href{https://wipolex.wipo.int/en/main/home}{Wipolex} \\ 
\hline
Portuguese & Judgments & 759 & 20590 & 6 & \href{https://wipolex.wipo.int/en/main/home}{Wipolex} \\
 & Laws & 1010 & 25947 & 3 &  \href{https://wipolex.wipo.int/en/main/home}{Wipolex}\\ 
\hline
German & Judgments & 21409 & 506009 & 131 & \href{https://github.com/sebischair/Legal-Sentence-Boundary-Detection}{\citet{Glaser}} \\
 & Laws & 20330 & 484816 & 13 & \href{https://github.com/sebischair/Legal-Sentence-Boundary-Detection}{\citet{Glaser}} \\ 
\hline
English & Judgments & 25899 & 712433 & 80 & \href{https://github.com/jsavelka/sbd_adjudicatory_dec}{\citet{Savelka}} \\ 
\hline
\textbf{Total} & Laws \& Judgments & 133568 & 3654145 & 906 &  \\
\bottomrule
\end{tabular}
\end{table*}

\subsection{Annotation}
\label{ss:Annotation}

The human annotator was tasked with correcting the sentence-spans predicted by an automatic \ac{SBD} system\footnote{https://github.com/jsavelka/luima\_sbd} \citep{Savelka} based on \ac{CRF}, which was trained on data annotated using annotation guidelines by \citet{Savelka}. This helped improve the quality and consistency of our annotations. Furthermore, a practical rule set, heavily influenced by the aforementioned guidelines, was utilized to aid the annotator in the annotation process, reducing the complexity of the task and helped provide dependable and well-founded data. The rule set is outlined in Section \ref{sss:legalSentenceStructure}, containing the most important sentence structures followed by an example.

The documents were annotated using Prodigy (\url{https://prodi.gy/}). Because Prodigy requires pre-tokenized text, a customized tokenizer was applied to the input text, further described in Section~\ref{ss:Tokenizer}.
The decision to annotate the full sentence-span, in lieu of just the first and last token in the sequence, was made to incentivize the annotator to read the text instead of skimming it for sentence-terminating characters.
To make the annotation easier, laws were split into smaller chunks with one to three articles per chunk, while judgments were only split, if they surpassed \textasciitilde{}15000 characters since Prodigy was unable to handle longer documents.

\subsubsection{Legal Sentence Structures}
\label{sss:legalSentenceStructure}
In this section, we briefly describe the most important sentence structures in legal text, heavily influenced by \citet{Savelka}, followed by an example in French.
\begin{description}
    \item[Standard Sentence] have subject, object and verb in the correct order and the last token in the sequence is a sentence-terminating character.
    \begin{itemize}
        \item \textit{Il s'est établi comme ingénieur indépendant.}
    \end{itemize}
    
    \item[Linguistically Transformed Sentence] are similar to a standard sentence, but slight transformations such as changes to the word order are applied.
    \begin{itemize}
        \item\textit{Tout porte à croire, en réalité, qu'elle est condamnée au surendettement, puis à la faillite.}
    \end{itemize}
    
    \item[Headlines] determine the structure of the text and show relatedness between parts of the document and therefore convey important information about the overall structure of the text.
    \begin{itemize}
        \item\textit{Considérant en fait et en droit}
        \item\textit{PAR CES MOTIFS}
        \item\textit{DÉCLARATION}
    \end{itemize}
    
    \item[Data fields] provide the name and data of a field. This is annotated as a sentence, as for example in English "Civil Chamber: Madrid" has a similar meaning to "The civil chambers are located in Madrid".
    \begin{itemize}
        \item \textit{Numéro d'appel: 1231/2015}
    \end{itemize}
    
    \item[Parentheses] appear frequently in legal text, often combined with citations. We annotate parentheses with the sentence they belong to. Sequences inside the parentheses are not annotated separately, as seen in the following example, containing a single sentence: 
    \begin{itemize}
        \item \textit{Ce dernier étant domicilié à l'étranger, il ne peut en effet prétendre à des mesures de réadaptation (art. 8a. 1er paragraphe. Convention de sécurité sociale entre la Suisse et la Yougoslavie du 8 juin 1962).}
    \end{itemize}

    \item[Colons] should not be annotated as a sentence-terminating character, unless the colon is immediately followed by a newline. The reasoning here is that a sequence ending in a colon followed by a line break usually introduce a list or block quote, which should be annotated separately to the introductory sentence.
    
    \item[Lists] are annotated differently depending on its type. For lists with incomplete sentences as list items, often ended with a semi-colon, the whole list is annotated as a sentence. The following example consists of 2 sentences, the introductory sentence to the colon and 1° to the period.
    \begin{itemize}
        \item \textit{Au cours du délai fixé par la juridiction pour accomplir un travail d'intérêt général, le condamné doit satisfaire aux mesures de contrôle suivantes:\\
        1° Répondre aux convocations du juge de l'application des peines;\\
        2° (...) une affection dangereuse pour les autres travailleurs.}
    \end{itemize}
    However, if the list items themselves are sentences, the list number (or letter) and items are both annotated as one sentence each, the reason being that they express separate thoughts. In the example below we have 3 sentences (introductory, list number, list item).
    \begin{itemize}
        \item \textit{Considérant en droit:\\1.- En instance fédérale, peut seul être examiné le point de savoir si la commission de recours a exigé à bon droit de la recourante une avance de frais de 500 fr. pour la procédure de recours de première instance.}
    \end{itemize}
    
    \item[Ellipses] are used to indicate when part of a sentence or part of the document are left out. 
    The following example shows the use cases for ellipses. The first ellipsis is annotated separately, as it indicates sentences that are missing. The second ellipses indicates, that part of that single sentence was left out and is therefore not annotated separately.
    \begin{itemize}
        \item \textit{(...) La faute de X. est d'une exceptionnelle gravité tant les faits qui lui sont reprochés (...), commis avec une certaine froideur sont insoutenables et comportent un caractère insupportable pour les victimes.}
    \end{itemize} 

    \item[Footnotes / Endnotes] convey additional information to the reader. Indicators for end- and footnotes such as numbers or letters should always be annotated as being inside the sentence span, even if they occur after the sentence-terminating character. As an example, the sequence below is just one sentence, with "(2)" as the indicator:
    \begin{itemize}
        \item La loi ne dispose que pour l'avenir; elle n'a point d'effet rétroactif. (2)
    \end{itemize}
    Furthermore, endnotes appearing as numbered lists, should be annotated as following the guidelines for lists. In the example below, (2) is one sentence, followed by a normal sentence:
    \begin{itemize}
        \item (2) Le remplacement des membres du Parlement a lieu conformément aux dispositions de l'article 25.
    \end{itemize}
    
\end{description}

%move to section 5?
\subsection{Tokenizer}
\label{ss:Tokenizer}
We implemented an aggressive tokenizer based on Regex to segment text into tokens, also employed in other research \citep{Glaser, Savelka}. This tokenizer was utilized for all languages. Words, numbers and special characters such as newlines and whitespace are separated into individual sequences. This was done to ensure no information (e.g., a line break indicating a sentence boundary), vital to the \ac{SBD} process, was lost. An example is showcased below; tokenized whitespace is left out for clarity's sake:
\begin{itemize}
    \item \textit{D.\_ est entré à l'école le 16 juillet 1979.}
    \item \textit{D | . | \_ | est | entré | à | l | ' | école | le | 16 | juillet | 1979 | . }
\end{itemize}

\section{Experimental Setup}
\label{ch:ExDesignEval}
We conducted a series of experiments to answer our research questions posed in Section \ref{sec:research_questions}. Firstly, we compared selected existing models to establish a baseline performance. Secondly, we trained and evaluated various monolingual and multilingual models based on \ac{CRF}, BiLSTM-CRF, and transformers, comparing them to baselines. Lastly, we evaluated the multilingual models' performance on unseen data in a zero-shot experiment.

\subsection{Baseline Systems}
We conducted a thorough evaluation of several widely used systems utilizing various technologies, including CoreNLP, NLTK, Stanza, and Spacy, which served as our baselines. In the following section, we will provide a detailed description of each system.

\subsubsection{NLTK}
A fully unsupervised \ac{SBD} system created by \citet{Kiss}. The main thought behind the system is that most falsely predicted sentence boundaries stem from periods after abbreviations. The system therefore discovers abbreviations by looking at the length, the collocational bond, internal periods and occurrences of abbreviations without an ending period of each token in the text. We test a pre-trained model as well as a model trained on our data.

\subsubsection{CoreNLP}
A rule-based system from the Stanford CoreNLP toolkit \citep{CorenlpManning}, which predicts sentence boundaries based on events like periods, question marks, or exclamation marks.

\subsubsection{Stanza}
A multilingual system based on a \ac{BiLSTM} model \citep{Qi.etal2020a}. We only use the first part of its \ac{NLP} pipeline, the tokenizer. It addresses tokenization and sentence splitting jointly, treating it as a character sequence tagging problem, predicting if a character is the end of a token or sentence.

\subsubsection{Spacy}
A multilingual system \citep{spacy} with pre-trained models using technologies like \ac{CNN} and transformers. For our purposes, only the tokenizer and sentence splitter were used.

\subsection{Our Models}
Following the works presented in Section \ref{ch:RelatedWork}, we chose to test models based on \acp{CRF}, BiLSTM-CRFs and transformers. We further describe these models in the following subsections. For testing, we trained\footnote{GPU: NVIDIA GeForce RTX 3060 TI, CPU: Intel Core i5-8600K CPU @ 3.60GHz} and evaluated monolingual models for each language as well as multilingual models using all languages except Portuguese, once for laws, once for judgments and both types together.

\subsubsection{Conditional Random Fields}
\label{sss:CRFExDesign}
%The tokenizer described in Subsection \ref{ss:Tokenizer} was used to tokenize the input text, including whitespaces. Each token is then translated into a list of relatively simple features, which represent the token. Additionally, for each token we add the features of tokens in a pre-defined window around the token; the size of each window is variable for each feature. Our chosen features and window-sizes are inspired by \citet{Glaser} and \citet{Savelka}, further described in Table \ref{tab:crf-features}. 
%We used the "BILOU" labeling system to label the input data, following \citet{Lin.etal2020}.
%"B-Sentence" and "L-Sentence" denote the first and last token in the sequence, while tokens inside and outside the sequence are denoted by "I-Sentence" and "O" respectively. "U" is for sequences consisting of one token.

The tokenizer in Section \ref{ss:Tokenizer} tokenized input text, including whitespaces. Each token was translated into a list of simple features representing the token, and the features of tokens within a pre-defined window around the token were added. Window sizes for each feature varied, inspired by \citet{Glaser} and \citet{Savelka}, as shown in Table \ref{tab:crf-features}. We labeled input data using the "BILOU" system following \citet{Lin.etal2020}.

\begin{table}[ht!]
\centering
\caption{Description of CRF-Features}
\label{tab:crf-features}
\begin{tabular}{lp{4.5cm}r} 
\toprule
Feature & Description & Window \\ 
\midrule
Special & Each token is categorized using the following translation: Sentence-terminating tokens as "End", opening and closing parentheses as "Open" and "Close" respectively, newline characters as "Newline", abbreviation characters as "Abbr" and the rest as "No". & 10 \\ 
\hline
Lowercase & The token in lowercase. & 7 \\ 
\hline
Length & The length of the token. & 7 \\ 
\hline
Signature & Each character is represented using the following translation: Lower case and upper case character are rewritten as "c" and "C" respectively, digits are written as "N" and special characters as "S". & 5 \\ 
\hline
Lower & Whether the first character is lower case. & 3 \\ 
\hline
Upper & Whether the first character is upper case. & 3 \\ 
\hline
Digit & Whether the token is a digit. & 3 \\
\bottomrule
\end{tabular}
\end{table}

For training our \ac{CRF} models, we used the python-crfsuite\footnote{https://pypi.org/project/python-crfsuite/} implementation. We trained each model for 100 iterations, with regularization parameters 1 and 1e$^{-3}$ for C1 and C2, L-BFGS as the algorithm, and including all possible feature transitions.

\subsubsection{Bidirectional LSTM - CRF}
A \ac{BiLSTM} connects two \acp{LSTM} with opposite directions to the same output, allowing it to capture information from past and future states at the same time. The outputs of each \ac{LSTM} are concatenated into a representation of each input token. For a \ac{BiLSTM}-\ac{CRF} model, a \ac{CRF} layer is connected to the output of the \ac{BiLSTM} network, using the aforementioned representation as features to predict the final label.

We utilized the Bi-LSTM-CRF\footnote{https://github.com/jidasheng/bi-lstm-crf} library to train our models.
We used a word embedding dimension of 128, hidden dimension of 256 and a maximum sequence length of 512. The batch-size was 16 with a learning rate of 0.01 and a weight decay of 0.0001. We trained each model for 8 epochs and saved the model with the smallest validation loss. 
We extracted word embeddings for training from our documents.
To label the training data, we utilized the "BILOU" labeling system described in Section \ref{sss:CRFExDesign}.
For training, gold sentences were put together into batches with a token-limit of 512 to simulate longer paragraphs.

\subsubsection{Transformer}
\label{sss:Transformer}
Transformers are a type of \ac{NN} that utilizes self-attention mechanisms to weigh the importance of difference parts of the input when making predictions. Transformer models such as BERT use a multi-layer encoder \citep{Vaswani.etal2017} to pre-train deep bidirectional representations by jointly conditioning on both left and right context across all layers \cite{Devlin.etal2019}. Thus, we can fine-tune transformer models to the SBD task by adding an additional output layer.
In our case we used a pre-trained model\footnote{https://huggingface.co/distilbert-base-multilingual-cased} based on DistilBERT \cite{Distil}, a smaller, more lightweight version of BERT, for all languages on our SBD task.\footnote{For efficiency, we used a smaller model; a bigger model is advisable for future work.}
We trained the models using PyTorch\footnote{https://pytorch.org/} and Accelerate\footnote{https://huggingface.co/docs/accelerate/index} with the Adam optimizer for 5 epochs with a batch-size of 8 and learning rate of 2e$^{-5}$.

A limitation of DistilBERT is the input length limit of 512 tokens because the runtime of the self-attention mechanism scales quadratically with the sequence length. This issue is exacerbated, since DistilBERT relies on a WordPiece Tokenizer \cite{Song.etal2020}, splitting the text into subwords resulting in a higher token count per sequence.
Thus, to get around the 512 token-limit, each document was split into sentences using the gold annotation. Each consecutive sentence was added to a collection until the total length was as close to the token-limit as possible. Next, the model predicted the sentence boundaries for each collection. Sentences longer than 512 tokens were truncated.\footnote{This led to some wrongly predicted sentence boundaries, however this only occurred a few times and is therefore insignificant to the overall score.} An obvious downside to this solution is that the input text already has to be split into sentences or short sections, making it difficult to apply BERT models to unknown text.
%DistilBERT has a limitation on input length, with a 512-token limit due to the quadratic scaling of the self-attention mechanism's runtime with sequence length. To address this issue, documents were split into sentences and grouped together to stay within the token limit. Sentence boundaries were predicted for each group, and longer sentences were truncated.\footnote{This led to some wrongly predicted sentence boundaries, however this only occurred a few times and is therefore insignificant to the overall score.} While this approach may lead to some incorrect predictions, it did not significantly affect the overall score. However, this method requires text to be pre-split into sentences or short sections, making it difficult to apply BERT models to unknown text.

For future work, it would be interesting to see, whether it is feasible to chain \ac{SBD} models (i.e., first, apply a \ac{CRF} model on the input text to split the text into sections smaller than 512 tokens and second apply a transformer based model). Another solution might be using pre-trained transformer models that support longer input text utilizing an attention mechanism scaling linearly with sequence length, such as Longformers \cite{Beltagy.etal2020}.\footnote{Unfortunately, to the best of our knowledge, so far there do not exist multilingually pretrained efficient transformer models.}

\begin{table*}
\centering
\caption{Mean (\textpm std) F1 Score of baseline and multilingual models on all languages and the Portuguese zero-shot experiment. Best scores are in bold.}
\label{tab:baselineMultilingual}
\resizebox{\textwidth}{!}{
\begin{tabular}{l|cc|cc|cc|cc|cc|cc|cc|cc|cc|cc|cc} 
\toprule
\multicolumn{1}{r|}{Language} & \multicolumn{4}{c|}{French} & \multicolumn{4}{c|}{Spanish} & \multicolumn{4}{c|}{Italian} & \multicolumn{2}{c|}{English} & \multicolumn{4}{c|}{German} & \multicolumn{4}{c}{Portuguese (Zero-shot)} \\
\hline
\multicolumn{1}{r|}{Type} & \multicolumn{2}{c|}{Judg.} & \multicolumn{2}{c|}{Laws} & \multicolumn{2}{c|}{Judg.} & \multicolumn{2}{c|}{Laws} & \multicolumn{2}{c|}{Judg.} & \multicolumn{2}{c|}{Laws} & \multicolumn{2}{c|}{Judg.} & \multicolumn{2}{c|}{Judg.} & \multicolumn{2}{c|}{Laws} & \multicolumn{2}{c|}{Judg.} & \multicolumn{2}{c}{Laws} \\
% & \multicolumn{2}{c|}{mean\textpm std} & \multicolumn{2}{c|}{mean\textpm std} & \multicolumn{2}{c|}{mean\textpm std} & \multicolumn{2}{c|}{mean\textpm std} & \multicolumn{2}{c|}{mean\textpm std} & \multicolumn{2}{c|}{mean\textpm std} & \multicolumn{2}{c|}{mean\textpm std} & \multicolumn{2}{c|}{mean\textpm std} & \multicolumn{2}{c|}{mean\textpm std} & \multicolumn{2}{c|}{mean\textpm std} & \multicolumn{2}{c}{mean\textpm std} \\
Model &  &  &  &  &  &  &  &  &  &  &  &  &  &  &  &  &  &  &  &  &  &  \\ 
\midrule
CoreNLP & \multicolumn{2}{c|}{74.7} & \multicolumn{2}{c|}{76.7} & \multicolumn{2}{c|}{71.4~} & \multicolumn{2}{c|}{89.0} & \multicolumn{2}{c|}{79.8} & \multicolumn{2}{c|}{75.6} & \multicolumn{2}{c|}{81.7} & \multicolumn{2}{c|}{69.0} & \multicolumn{2}{c|}{64.0} & \multicolumn{2}{c|}{-} & \multicolumn{2}{c}{-} \\
NLTK & \multicolumn{2}{c|}{72.5} & \multicolumn{2}{c|}{75.8} & \multicolumn{2}{c|}{70.2} & \multicolumn{2}{c|}{89.2} & \multicolumn{2}{c|}{72.3} & \multicolumn{2}{c|}{66.3} & \multicolumn{2}{c|}{77.2} & \multicolumn{2}{c|}{72.3} & \multicolumn{2}{c|}{73.8} & \multicolumn{2}{c|}{64.9} & \multicolumn{2}{c}{57.0} \\
NLTK-train & \multicolumn{2}{c|}{82.9} & \multicolumn{2}{c|}{75.8} & \multicolumn{2}{c|}{72.1} & \multicolumn{2}{c|}{81.6} & \multicolumn{2}{c|}{84.8} & \multicolumn{2}{c|}{77.5} & \multicolumn{2}{c|}{84.9} & \multicolumn{2}{c|}{74.2} & \multicolumn{2}{c|}{73.5} & \multicolumn{2}{c|}{71.7} & \multicolumn{2}{c}{64.3} \\
Spacy & \multicolumn{2}{c|}{86.6} & \multicolumn{2}{c|}{67.2} & \multicolumn{2}{c|}{60.0} & \multicolumn{2}{c|}{70.3} & \multicolumn{2}{c|}{73.9} & \multicolumn{2}{c|}{73.7} & \multicolumn{2}{c|}{79.7} & \multicolumn{2}{c|}{87.5} & \multicolumn{2}{c|}{67.0} & \multicolumn{2}{c|}{59.0} & \multicolumn{2}{c}{77.7} \\
Stanza & \multicolumn{2}{c|}{81.9} & \multicolumn{2}{c|}{81.0} & \multicolumn{2}{c|}{83.2} & \multicolumn{2}{c|}{90.2} & \multicolumn{2}{c|}{85.7} & \multicolumn{2}{c|}{87.4} & \multicolumn{2}{c|}{92.3} & \multicolumn{2}{c|}{72.6} & \multicolumn{2}{c|}{64.7} & \multicolumn{2}{c|}{88.6} & \multicolumn{2}{c}{73.4} \\ 
\hline
CRF & \multicolumn{2}{c|}{97.8} & \multicolumn{2}{c|}{98.1} & \multicolumn{2}{c|}{94.8} & \multicolumn{2}{c|}{98.9} & \multicolumn{2}{c|}{97.3} & \multicolumn{2}{c|}{97.7} & \multicolumn{2}{c|}{95.1} & \multicolumn{2}{c|}{95.2} & \multicolumn{2}{c|}{91.6} & \multicolumn{2}{c|}{90.2} & \multicolumn{2}{c}{78.6} \\
BiLSTM-CRF & \multicolumn{2}{c|}{97.6\textpm0.3} & \multicolumn{2}{c|}{\textbf{98.5}\textpm0.2} & \multicolumn{2}{c|}{97.3\textpm0.1} & \multicolumn{2}{c|}{\textbf{99.3}\textpm0.2} & \multicolumn{2}{c|}{97.8\textpm0.1} & \multicolumn{2}{c|}{\textbf{99.2}\textpm0.1} & \multicolumn{2}{c|}{95.4\textpm0.3} & \multicolumn{2}{c|}{\textbf{97.2}\textpm0.2} & \multicolumn{2}{c|}{97.5\textpm0.5} & \multicolumn{2}{c|}{93.0\textpm0.6} & \multicolumn{2}{c}{73.2\textpm3.3} \\
Transformer & \multicolumn{2}{c|}{\textbf{98.3}\textpm0.1} & \multicolumn{2}{c|}{98.1\textpm0.2} & \multicolumn{2}{c|}{\textbf{97.8}\textpm0.1} & \multicolumn{2}{c|}{99.0\textpm0.0} & \multicolumn{2}{c|}{\textbf{98.3}\textpm0.1} & \multicolumn{2}{c|}{99.1\textpm0.1} & \multicolumn{2}{c|}{\textbf{97.0}\textpm0.1} & \multicolumn{2}{c|}{92.9\textpm0.2} & \multicolumn{2}{c|}{\textbf{97.6}\textpm0.1} & \multicolumn{2}{c|}{\textbf{93.6}\textpm0.3} & \multicolumn{2}{c}{\textbf{91.3}\textpm1.1} \\
\bottomrule
\end{tabular}}
\end{table*}

\subsection{Evaluation}
A characteristic of the SBD task is the inherent imbalance towards non-sentence boundary labels, as each sentence can at most have two sentence boundaries. Thus, to more accurately score our models, we used commonly utilized measures to evaluate our models - \ac{P}, \ac{R} and \ac{F1}. Although the SBD task is not yet solved in specialized domains, it is comparatively easier than other NLP tasks such as Questions Answering or Summarization. Because SBD is a pre-processing task, it is necessary to achieve higher scores to prohibit the propagation of errors into downstream tasks. Thus, we expect that state-of-the-art SBD models exhibit F1-scores in the high nineties to be useful in practice. 

For the evaluation process, we let models predict the sentence spans of every document. These annotated spans are tokenized by our tokenizer (Section \ref{ss:Tokenizer}). Each token is then assigned a binary value, depending on whether it was a sentence boundary or not. This decouples the predicted sentence spans or boundaries from the tokenizer used, as the tokenizer of some models might designate a slightly different token as the first or last in a sentence, further described in the following example in French: \textit{"C'est en outre ...".}
While our tokenizer would designate "C" as the first token in the sequence, a different tokenizer might designate "C'" or even "C'est".
This would lead to a wrongly predicted sentence boundary when compared to the gold annotations, although the prediction was actually correct.

%The true and predicted labels are gathered for every document type and then compared using Scikit-Learn\footnote{https://scikit-learn.org/stable/}, calculating the binary F1-Score. Finally, the scores are averaged for each subset, with Criminal Code, Civil Code and Constitution being reported under "Laws" and the various court decisions under "Judgments".
True and predicted labels for each document type are compared using Scikit-Learn to calculate binary F1-Scores. Scores are averaged for subsets: "Laws" encompass Criminal Code, Civil Code, and Constitution; "Judgments" include various court decisions.

We trained each \ac{CRF} model once and the BiLSTM-CRF and transformer models 5 times with random seeds, reporting the mean performance including standard deviation. If not specified differently, reported values are binary \ac{F1}-scores.

\section{Results}
\label{ch:Results}

\subsection{Baseline Models}
\label{ss:Baseline}
The performance of baseline models in Section \ref{ch:ExDesignEval} on each language in our dataset is summarized in the upper section of Table~\ref{tab:baselineMultilingual}. 
%A more detailed account for each language and baseline model is shown in Appendix \ref{ss:baselineModelsApp}.

The results for the baseline models are clearly lower than the reported scores for user-generated content by \citet{Read}, supporting the hypothesis that the performance of out-of-the-box models is subpar on legal data for all tested languages. The difference in performance could be explained in one part by the special sentence structures presented in Section \ref{ss:Annotation}, while the challenging nature of legal text accounts for another part.

Of interest is the gap between NLTK and NLTK-train in most languages, as training NLTK improves its ability to recognize and correctly predict abbreviations. This showcases that abbreviations are one part of the challenging nature of legal texts.
To note here is that Spacy uses a slightly different notion of a sentence compared to the other models: Usually, when two sentences are separated by a newline character, the newline character would not be part of any sentence span, however Spacy would include it in the span of the second sentence. This leads to a false prediction, even though Spacy correctly recognized that there are two sentences. Therefore, the scores Spacy achieves are lower than expected.

%-----------------------------------------------------------------------------------------------------------%

\subsection{Monolingual Models}
\label{ss:monoModels}
We report the performance of our trained monolingual models in Table \ref{tab:monolinguals}. Each model was trained and tested on the same language. 
%A more detailed account is shown in Appendix \ref{ss:trainedModelsApp}.

We observe that each model's performance, when applied to their training subset, reaches high nineties for almost all languages, significantly improving over the baseline models from Section \ref{ss:Baseline} and comparable to reported SBD system performance on English news articles \citep{Read}. Our models also perform similarly to the reported performance of \acp{CRF} and \acp{CNN} on English \citep{Savelka, Sanchez}, as well as \acp{CRF} and \acp{NN} on German datasets \citep{Glaser}.

Comparing the performance of the models when trained on one subset and evaluated on the other unseen set, i.e. a zero-shot experiment, the transformer model outperforms \ac{CRF} and BiLSTM-CRF on most languages, dropping down to 81.8\% on the Italian dataset, comparable to the best baseline models, when trained on judgements and evaluated on laws. 
Unsurprisingly, the models' performance in the zero-shot experiment is almost always lower than the performance on the subset they were trained on. This gap can be explained by the large difference of writing and formatting styles between judgements and laws, with the transformer model being the best at generalizing knowledge between the two subsets. We further hypothesize that it was easier for the models to generalize their knowledge to different domains, when being trained on judgements, than when being trained on laws, resulting in higher scores on unseen data. One factor here might be that legal text in judgements contain a higher variety of different sentence structures, while laws usually reuse the same structures.

The CRF and BiLSTM-CRF model showcase especially poor performance on the Spanish dataset when trained on laws and evaluated on judgements, with scores down to 43.4\% and 54.3\%. We hypothesize that both models possess a worse ability to generalize to different domains compared to transformer models. 

To conclude, while training on both laws and judgments together not always produces the absolute best performance, it is most robust and does not result in performance degradation. 

\begin{table*}
\centering
\caption{Mean (\textpm std) F1 Score of monolingual models on their respective language. Best scores are in bold.}
\label{tab:monolinguals}
\resizebox{\textwidth}{!}{
\begin{tabular}{l|l|c|c|c|c|c|c|c|c|c} 
\toprule
\multicolumn{1}{l}{} & \multicolumn{1}{r|}{Language} & \multicolumn{2}{c|}{French} & \multicolumn{2}{c|}{Spanish} & \multicolumn{2}{c|}{Italian} & English & \multicolumn{2}{c}{German} \\
\hline
\multicolumn{1}{l}{} & \multicolumn{1}{r|}{Type} & Judg. & Laws & Judg. & Laws & Judg. & Laws & Judg. & Judg. & Laws \\
%\multicolumn{1}{l}{} &  & mean\textpm std & mean\textpm std & mean\textpm std & mean\textpm std & mean\textpm std & mean\textpm std & mean\textpm std & mean\textpm std & mean\textpm std \\
Model & Trained on &  &  &  &  &  &  &  &  &  \\ 
\midrule
\multirow{3}{*}{CRF} & Judg. & 97.9 & 73.2 & 97.0 & 98.3 & \textbf{98.5} & 95.6 & 96.8 & 97.8 & 76.5 \\
 & Laws & 78.5 & \textbf{98.8} & 54.3 & \textbf{99.6} & 88.6 & \textbf{99.6} & - & 75.8 & \textbf{97.7} \\
 & Laws + Judg. & 97.8 & \textbf{98.8} & 97.0 & 99.5 & 98.3 & 99.5 & - & 97.2 & 97.2 \\ 
\hline
\multirow{3}{*}{BiLSTM-CRF} & Judg. & 97.3\textpm0.3 & 56.7\textpm3.0 & 94.7\textpm0.5 & 92.1\textpm0.9 & 95.9\textpm0.3 & 71.3\textpm2.2 & \textbf{97.3}\textpm0.4 & 97.0\textpm0.3 & 76.9\textpm0.4 \\
 & Laws & 66.1\textpm4.2 & 97.9\textpm0.2 & 43.4\textpm6.8 & 98.7\textpm0.2 & 74.1\textpm1.2 & 98.4\textpm0.2 & - & 71.9\textpm2.5 & 97.3\textpm0.3 \\
 & Laws + Judg. & 97.0\textpm0.4 & 98.1\textpm0.1 & 95.6\textpm0.5 & 98.9\textpm0.4 & 96.2\textpm0.2 & 98.2\textpm0.1 & - & 97.2\textpm0.2 & 97.6\textpm0.2 \\ 
\hline
\multirow{3}{*}{Transformer} & Judg. & 98.2\textpm0.1 & 84.7\textpm1.2 & 96.9\textpm0.2 & 96.9\textpm0.4 & 97.8\textpm0.2 & 81.8\textpm0.9 & 96.5\textpm0.1 & 98.0\textpm0.2 & 87.2\textpm0.4 \\
 & Laws & 92.4\textpm0.5 & 97.6\textpm0.4 & 89.5\textpm0.6 & 97.1\textpm3.7 & 89.4\textpm0.7 & 98.8\textpm0.5 & - & 89.4\textpm0.5 & 97.4\textpm0.1 \\
 & Laws + Judg. & \textbf{98.4}\textpm0.1 & 98.2\textpm0.2 & \textbf{97.3}\textpm0.1 & 99.0\textpm0.1 & 97.1\textpm0.3 & 99.1\textpm0.1 & - & \textbf{98.3}\textpm0.1 & 97.5\textpm0.2 \\
\bottomrule
\end{tabular}}
\end{table*}

%-----------------------------------------------------------------------------------------------------------%
\subsection{Multilingual Models}
\label{ss:multiModels}
The performance of our multilingual models trained on laws and judgements is reported in the lower section of Table \ref{tab:baselineMultilingual}. Each multilingual model was trained on all languages except Portuguese. 
%A more detailed report including multilingual models trained on all laws or all judgements separately is available in Appendix \ref{ss:multilingualApp}.

The multilingual models clearly outperform the baseline models by a large margin, with F1-scores up to 99.2\%. Both the BiLSTM-CRF and transformer models perform very well, with transformers performing slightly better on judgements and BiLSTM-CRFs on laws. The CRF model is close behind the other two, mostly reaching scores in the higher nineties.
Comparing the performance of the multilingual models to the monolingual models, showcases that there is no loss of performance when training on a much larger dataset, with multilingual models performing comparably or in case of the transformer and BiLSTM-CRF model even better than the monolingual models on each respective language.

\subsection{Zero-shot Experiment on Portuguese Data}
\label{ss:zeroshot}
%As an additional experiment, we evaluated the multilingual models from the previous section on unseen Portuguese data, reporting their F1-scores compared to the pre-trained Portuguese baseline in Table \ref{tab:baselineMultilingual}, the overall mean performance in Fig. \ref{fig:zeroB} and in more detail in Appendix \ref{ss:multilingualApp}. Note that we do not report an evaluation for CoreNLP on Portuguese as there was no pre-trained model available.

%As an additional, more challenging experiment, we evaluated the multilingual models from the previous section on unseen Portuguese data and compared them to the baseline. An overview can be found in Fig. \ref{fig:teaser}. More detailed results are shown in Table \ref{tab:baselineMultilingual}, showcasing the difference between the judgements and laws compared to the baseline. 
We conducted a more challenging experiment, evaluating multilingual models on Portuguese data, comparing them to the baseline. Figure \ref{fig:teaser} provides an overview, while Table \ref{tab:baselineMultilingual} details the differences in judgements and laws against the baseline.
%A full report, including Recall and Precision, is shown in Appendix \ref{ss:multilingualApp}. 

%For judgements, the performance is still adequate, reaching F1-scores between 90.2\% and 93.6\%, comparable to the reported performance on user-generated content \cite{Read}, outperforming most of the baseline models. However, for laws only the transformer model reaches the lower nineties, while for CRF and BiLSTM-CRF, the scores drop to 78.6\% and 73.2\% respectively, comparable to our usual baselines values. We hypothesize that the large-scale multilingual pretraining of the transformer model makes it more robust towards distribution shifts, leading to better cross-lingual transfer to unseen languages compared to CRFs or BiLSTM-CRFs.
For judgements performance is adequate with F1-scores between 90.2\% and 93.6\%, comparable to user-generated content \cite{Read}, and outperforming most baselines. However, for laws, only the transformer model scores in the lower nineties, while CRF and BiLSTM-CRF drop to 78.6\% and 73.2\%, respectively, similar to our usual baseline values. The transformer model's large-scale multilingual pretraining likely makes it more robust to distribution shifts, leading to better cross-lingual transfer to unseen languages than CRFs or BiLSTM-CRFs.

%The difference between laws and judgements could be explained by the increased difficulty of the writing and formatting style in Portuguese law texts, also indicated by the lower than usual Portuguese baseline performance. Another reason for the lowered performance of the BiLSTM-CRF could be the lack of Portuguese word embeddings used in training, as we extracted the embeddings from our training data. For future research, it would be interesting to see whether adding Portuguese word embeddings or using even larger, multilingual embedding vocabularies during training would increase the performance of the BiLSTM-CRF models.
%A potential avenue for further improving the transformer models might be the utilization of larger pre-trained models for fine-tuning on the \ac{SBD} task, such as XLM-RoBERTa \citep{Conneau.etal2020}, which reports significant improvements on cross-lingual transfer tasks compared to BERT-based models. 

The difficulty of the writing and formatting style in Portuguese law texts could explain the difference between laws and judgements, indicated by lower than usual Portuguese baseline performance. BiLSTM-CRF's reduced performance could also result from the lack of Portuguese word embeddings used in training, as we only extracted embeddings from our training data. To improve BiLSTM-CRF models, future research could explore adding Portuguese word embeddings or using larger, multilingual embedding vocabularies during training. To improve transformer models, fine-tuning larger pre-trained models like XLM-RoBERTa \citep{Conneau.etal2020} on the \ac{SBD} task could be a potential avenue as they improve significantly in cross-lingual transfer compared to mBERT \cite{Devlin.etal2019} or DistilBERT \cite{Distil} models.

\begin{table}
\caption{Mean F1 Score of monolingual and multilingual models on unseen Portuguese data}
\label{tab:monoZero}
\resizebox{\columnwidth}{!}{
\begin{tabular}{l|c|c|c|c|c|c} 
\toprule
\multicolumn{1}{r|}{Model} & \multicolumn{2}{c|}{CRF} & \multicolumn{2}{c|}{BiLSTM-CRF} & \multicolumn{2}{c}{Transformer} \\ 
\hline
\multicolumn{1}{r|}{Type} & Judg. & Laws & Judg. & Laws & Judg. & Laws \\
Model Language &  &  &  &  &  &  \\ 
\midrule
French & 79.3 & 75.4 & 25.5 & 51.7 & 82.5 & 87.1 \\
Spanish & \textbf{91.5} & 79.4 & 80.3 & \textbf{73.5} & 88.0 & \textbf{94.0} \\
Italian & 81.8 & \textbf{83.3} & 12.6 & 64.8 & 70.0 & 73.7 \\
English & 90.6 & 72.1 & 80.6 & 62.4 & 87.6 & 89.9 \\
German & 59.0 & 25.2 & 43.6 & 30.3 & 79.9 & 71.1 \\ 
\hline
Multilingual & 90.2 & 78.6 & \textbf{93.0} & 73.2 & \textbf{93.6} & 91.3 \\
\bottomrule
\end{tabular}}
\end{table}

When evaluating the effectiveness of monolingual and multilingual models, trained on the entire monolingual dataset, on previously unseen Portuguese data (Table \ref{tab:monoZero}), we observe that the multilingual models outperform corresponding monolingual models in most languages, with Spanish being a notable exception. We hypothesize that the disparity in performance is due to close linguistic ties between Spanish and Portuguese, which enabled the Spanish monolingual models to excel in cross-lingual transfer. However, on other languages linguistically less close to Spanish, the multilingual model is expected to perform better than the monolingual ones. 

%-----------------------------------------------------------------------------------------------------------%
\subsection{Inference Time}
\label{ss:inf_times}

%The inference times of our multilingual models trained on both laws and judgements are reported in Table \ref{tab:inferenceTime}. We measured inference time thrice on a GPU (NVIDIA GeForce RTX 3060 TI) and CPU (Intel Core i5-8600K CPU @ 3.60GHz) and show the average. We do not report standard deviation since there were no significant outliers. Large improvements appear when measuring inference time on a GPU, especially for the transformer model. Note that \ac{CRF} uses sequential operations, not batch operations, so it does not benefit from GPU evaluation.

Table \ref{tab:inferenceTime} reports the inference times of our multilingual models trained on laws and judgments. We measured inference time three times on both a GPU (NVIDIA GeForce RTX 3060 TI) and a CPU (Intel Core i5-8600K CPU @ 3.60GHz), and show the average. We did not report standard deviation since there were no significant outliers. Notably, the transformer model saw significant improvements in inference time on a GPU. However, \ac{CRF} does not benefit from GPU evaluation as it uses sequential operations.

\begin{table}[ht]
\centering
\caption{Mean inference time in minutes (min), seconds (s), milliseconds (ms) for each multilingual model to predict the entire dataset of \textasciitilde{}130000 sentences and one sentence, measured on a GPU and CPU}
\label{tab:inferenceTime}
\resizebox{\columnwidth}{!}{
\begin{tabular}{l|r|c|r|c} 
\toprule
 & \multicolumn{2}{c|}{full dataset (\textasciitilde{}130000 sentences)} & \multicolumn{2}{c}{One sentence} \\ 
Model & \multicolumn{1}{c|}{CPU} & GPU & \multicolumn{1}{c|}{CPU} & GPU \\ 
\midrule
CRF & 11 min 57 sec & - & \textasciitilde{}5.37 ms & - \\
BiLSTM-CRF & 10 min \hspace{0.8mm} 6 sec & \multicolumn{1}{r|}{9 min 23 sec} & \textasciitilde{}4.54 ms & \multicolumn{1}{r}{\textasciitilde{}4.21 ms} \\
Transformer & 34 min 26 sec & \multicolumn{1}{r|}{9 min 18 sec} & \textasciitilde{}15.47 ms & \multicolumn{1}{r}{\textasciitilde{}4.18~ ms} \\
\bottomrule
\end{tabular}}
\end{table}
Considering the results presented in Sections \ref{ss:monoModels}, \ref{ss:multiModels} and \ref{ss:zeroshot}, inference times and ease of use, a recommendation for the multilingual transformer model can be made for most cases, as long as a GPU is available for inference. For language specific tasks or tasks requiring longer input texts, we recommend the CRF models for the respective language, although they have a longer setup time compared to the BiLSTM-CRF and transformer model.
%Our transformer models can be found on 
%\href{https://huggingface.co/tbrugger}{HuggingFace}
%HuggingFace
%and our CRF and BiLSTM-CRF models as well as the code on 
%\href{https://github.com/tobiasbrugger/MultiLegalSBD}{GitHub}. !!swap to rcds
%GitHub.

\subsection{Error Analysis}
\label{sec:error_analysis}

We inspected random samples -- two thirds of the Portuguese dataset (8 judgements, 20 laws) -- predicted by the multilingual transformer model for the zero-shot experiment on Portuguese texts. We selected the multilingual transformer following our recommendation in Section \ref{ss:inf_times}, and the Portuguese dataset because the model already performed very well on the other datasets.

Standard sentence boundaries are rarely missed and the model performs adequately in that regard; yet, we identified a few sources of common mistakes. We discuss examples with |T| and |P| indicating true and predicted sentence boundaries, respectively. 
Many errors stem from citations and parentheses as shown in the example below:

\begin{itemize}
    \item (Bittar, Carlos Alberto. \emph{|P|} Direito de autor. \emph{|P|} Rio de Janeiro: Forense Universitária, 2001, p. 143) \emph{|T| |P|}
\end{itemize}
In this example, we have a citation sentence with periods being wrongly predicted as sentence boundaries inside the citation.

Another source of errors are datafields and headlines, since there is often little indication e.g., a sentence-terminating character, for the model to recognize it as such:
\begin{enumerate}
    \item RELATOR: MINISTRO SIDNEI BENETI \emph{|T|}
    \item ACÓRDÃO \emph{|T|}
\end{enumerate}
% add list / ellipses
The model failed to predict a sentence boundary at the end of both sequences. 
The errors showcased in the examples above mainly stem from our particularly defined sentence structures (Section \ref{sss:legalSentenceStructure}) as well as the challenging nature of the legal SBD task.

Another set of errors were caused by the different formatting styles and words used in the Portuguese language, unknown to the model, such as:
\begin{enumerate}
    \item A Turma, por unanimidade, deu provimento ao recurso especial, nos termos do voto do(a) Sr(a). \emph{|P|} Ministro(a) Relator(a). \emph{|T| |P|}
    \item Exmos. \emph{|P|} Desembargadores MAURÍCIO PESSOA (Presidente), CLAUDIO GODOY E GRAVA BRAZIL. \emph{|T| |P|}
\end{enumerate}
In (1), we have the abbreviation "Sr(a)", which the model did not recognise as such, thus marking the period as a sentence boundary. A similar mistake is shown in (2), with the abbreviation "Exmos".

\subsection{Limitations}

% Only two language groups (Germanic, Italic) with large lexical overlap: cross-lingual transfer "easy"
Due to the language skills of our annotator, we only annotated data from two language groups (Germanic and Italic). Therefore, our languages have high lexical overlap, making cross-lingual transfer comparatively easy. Future work may investigate legal text from additional diverse language groups to build systems even more robust towards language distribution shifts.

% Annotator is native German speaker, use annotators with native skills
The annotator is a native German speaker, with intermediate French language skills. Due to the similarity of Italian, Spanish, and Portuguese to French, and because the SBD task is largely structural, the annotations were possible. However, having the annotations performed by a native speaker in the respective languages may further increase annotation quality. On the other hand, having one annotator (as done in our case) annotate the entire dataset, enables more consistency across languages. 

% Have another annotator to validate annotations
Because of financial limitations, we performed the annotations using only one annotator. Having a second annotator validate the annotations may further increase annotation quality.

% Quantitative error analysis would be great 
Augmenting the qualitative error analysis from Section~\ref{sec:error_analysis} quantitatively may provide more concrete and actionable evidence for improving the systems further. To achieve this, a more detailed annotation of the sentence type would be helpful, so statistics over the sentences can be computed to get quantitative results of the sentence types performing worst.
%To obtain concrete evidence for improving systems, augmenting the qualitative error analysis from Section~\ref{sec:error_analysis} quantitatively is recommended. This requires a more detailed annotation of sentence type to compute statistics and identify poorly performing sentence types.

% More hyperparameter tuning (not enough compute)

% Portuguese Dataset is small

\section{Conclusion and Future Work}
\label{ch:Conclusion}

\subsection{Answers to the Research Questions}

\begin{description}
    \item[RQ1:] \emph{What is the performance of existing SBD systems on legal data in French, Spanish, Italian, English, and German?}\\
    Existing SBD systems are subpar in all tested languages, lower than reported scores by \citet{Read} on user-generated content, indicating that SBD is not solved in the legal domain.
    \item[RQ2:] \emph{To what extent can we improve upon this performance by training mono- and multilingual models based on CRF, BiLSTM-CRF and transformers?}\\
    %All monolingual models reach F1-scores in the higher nineties on all tested languages, demonstrating state-of-the-art performance, comparable to reported scores on news articles \citep{Read}. The multilingual models perform comparably to the monolingual models, illustrating the possibility of training with larger datasets. The transformer model showcases better cross-domain transfer compared to the CRF and BiLSTM-CRF models.
    The monolingual models achieved state-of-the-art F1-scores in the high nineties for all tested languages, comparable to reported scores on news articles \citep{Read}. The multilingual models performed similarly to monolingual models, demonstrating the potential of training with larger datasets. The transformer model exhibited superior cross-domain transfer compared to CRF and BiLSTM-CRF models.
    \item[RQ3:] \emph{What is the performance of the multilingual models on unseen Portuguese legal text, i.e., a zero-shot experiment?}\\
    The transformer models performs adequately on the judgements and laws subsets, reaching F1-scores in the lower nineties, demonstrating the best cross-lingual transfer, while the CRF and BiLSTM-CRF models perform decently around 90\% on judgements, but drop down to baseline values on the laws, most likely requiring additional optimization.
\end{description}

\subsection{Conclusion}
In this work, we curated and publicly released a diverse legal dataset with over 130'000 annotated sentences in 6 languages, enabling further research in the legal domain.
Using this dataset, we showed that existing SBD methods perform poorly on multilingual legal data, at most reaching F1-scores in the low nineties. We trained and evaluated mono- and multilingual \ac{CRF}, BiLSTM-CRF and transformer models, achieving binary F1-scores in the higher nineties on our dataset, demonstrating state-of-the art performance. For a more challenging task, we tested our multilingual models in a zero-shot experiment on unseen Portuguese data, with the transformer model reaching scores in the lower nineties, outperforming the baseline trained on Portuguese texts as well as the CRF and BiLSTM-CRF models by a large margin. We publicly release these models and the code for further use and research in the community.
%We showcased the sub-optimal performance of existing systems, state-of-the art performance of our trained mono- and multilingual models and that transformers have the best ability to generalize knowledge to other domains and languages, clearly outperforming CRF and BiLSTM-CRF. 

%We presented the sub-optimal performance of existing baseline SBD systems, when applied on legal documents in all tested languages. We trained and evaluated monolingual SBD models based on CRFs, BiLSTM-CRFs and transformers, outperforming the baseline by a large margin in all languages, achieving state-of-the-art performance. We presented results of multilingual SBD models based on the same technologies performing comparably or better than the monolingual models, showcasing their feasibility. We tested these multilingual models in a zero-shot experiment, applying them on previously unseen data. The transformer models outperformed the CRF and BiLSTM-CRF, showcasing a better cross-lingual knowledge transfer ability and being better suited to the task. We publicly release the models for further use and research in the community.
%write about downside e.g. weakened performance when applied to another subset

\subsection{Future Work}
Further improvement for all models might be achieved by pre-processing the input text more, e.g., replacing newlines with spaces, special characters with more widely used equivalent characters e.g., double quotes (“) with single quotes (").
Furthermore, thorough hyperparameter optimization tailored to the specific dataset could improve multilingual CRF and BiLSTM-CRF models.
Finally, transformer models may benefit from legal-oriented models \cite{chalkidis_legal-bert_2020, niklaus_budgetlongformer_2022, hua_legalrelectra_2022}, larger pre-trained models like BERT \cite{Devlin.etal2019}, or models designed for cross-lingual transfer tasks, like XLM-RoBERTa \cite{Conneau.etal2020}.

Augmenting the dataset with legal texts from multiple languages and documents from various sources like privacy policies and terms of service may improve multilingual models' performance, particularly in the zero-shot scenario. 
An interesting impact on the model performance could be observed if the sentence spans were labeled with their sentence structure type such as "Citation" (Section \ref{sss:legalSentenceStructure}) during training instead of being assigned a single label.

%Lastly, based on the assumption that the impressive cross-lingual transfer observed in the zero-shot experiment is partially due to the common origin of all languages in our study (refer to Section \ref{ch:Results}), it would be interesting to explore if this positive cross-lingual transfer applies to languages from a different language family, such as Hungarian.

An investigation into whether the positive cross-lingual transfer observed in their study also applies to languages from a different family, such as Hungarian. This assumption is based on the common origin of the languages studied, as mentioned in Section \ref{ch:Results}.

%%
%% The next two lines define the bibliography style to be used, and
%% the bibliography file.
\bibliographystyle{ACM-Reference-Format}
\bibliography{Bibliography,references}

%%% -*-BibTeX-*-
%%% Do NOT edit. File created by BibTeX with style
%%% ACM-Reference-Format-Journals [18-Jan-2012].

\begin{thebibliography}{34}

%%% ====================================================================
%%% NOTE TO THE USER: you can override these defaults by providing
%%% customized versions of any of these macros before the \bibliography
%%% command.  Each of them MUST provide its own final punctuation,
%%% except for \shownote{}, \showDOI{}, and \showURL{}.  The latter two
%%% do not use final punctuation, in order to avoid confusing it with
%%% the Web address.
%%%
%%% To suppress output of a particular field, define its macro to expand
%%% to an empty string, or better, \unskip, like this:
%%%
%%% \newcommand{\showDOI}[1]{\unskip}   % LaTeX syntax
%%%
%%% \def \showDOI #1{\unskip}           % plain TeX syntax
%%%
%%% ====================================================================

\ifx \showCODEN    \undefined \def \showCODEN     #1{\unskip}     \fi
\ifx \showDOI      \undefined \def \showDOI       #1{#1}\fi
\ifx \showISBNx    \undefined \def \showISBNx     #1{\unskip}     \fi
\ifx \showISBNxiii \undefined \def \showISBNxiii  #1{\unskip}     \fi
\ifx \showISSN     \undefined \def \showISSN      #1{\unskip}     \fi
\ifx \showLCCN     \undefined \def \showLCCN      #1{\unskip}     \fi
\ifx \shownote     \undefined \def \shownote      #1{#1}          \fi
\ifx \showarticletitle \undefined \def \showarticletitle #1{#1}   \fi
\ifx \showURL      \undefined \def \showURL       {\relax}        \fi
% The following commands are used for tagged output and should be
% invisible to TeX
\providecommand\bibfield[2]{#2}
\providecommand\bibinfo[2]{#2}
\providecommand\natexlab[1]{#1}
\providecommand\showeprint[2][]{arXiv:#2}

\bibitem[Aumiller et~al\mbox{.}(2022)]%
        {aumiller_eur-lex-sum_2022}
\bibfield{author}{\bibinfo{person}{Dennis Aumiller}, \bibinfo{person}{Ashish
  Chouhan}, {and} \bibinfo{person}{Michael Gertz}.}
  \bibinfo{year}{2022}\natexlab{}.
\newblock \bibinfo{title}{{EUR}-{Lex}-{Sum}: {A} {Multi}- and {Cross}-lingual
  {Dataset} for {Long}-form {Summarization} in the {Legal} {Domain}}.
\newblock
\newblock
\urldef\tempurl%
\url{http://arxiv.org/abs/2210.13448}
\showURL{%
\tempurl}
\newblock
\shownote{arXiv:2210.13448 [cs]}.


\bibitem[Beltagy et~al\mbox{.}(2020)]%
        {Beltagy.etal2020}
\bibfield{author}{\bibinfo{person}{Iz Beltagy}, \bibinfo{person}{Matthew~E.
  Peters}, {and} \bibinfo{person}{Arman Cohan}.}
  \bibinfo{year}{2020}\natexlab{}.
\newblock \bibinfo{title}{Longformer: {{The Long-Document Transformer}}}.
\newblock
\newblock
\showeprint[arxiv]{2004.05150}~[cs]


\bibitem[Chalkidis et~al\mbox{.}(2021)]%
        {chalkidis_multieurlex_2021}
\bibfield{author}{\bibinfo{person}{Ilias Chalkidis}, \bibinfo{person}{Manos
  Fergadiotis}, {and} \bibinfo{person}{Ion Androutsopoulos}.}
  \bibinfo{year}{2021}\natexlab{}.
\newblock \showarticletitle{{MultiEURLEX} -- {A} multi-lingual and multi-label
  legal document classification dataset for zero-shot cross-lingual transfer}.
\newblock \bibinfo{journal}{\emph{arXiv:2109.00904 [cs]}}
  (\bibinfo{date}{Sept.} \bibinfo{year}{2021}).
\newblock
\urldef\tempurl%
\url{http://arxiv.org/abs/2109.00904}
\showURL{%
\tempurl}
\newblock
\shownote{arXiv: 2109.00904}.


\bibitem[Chalkidis et~al\mbox{.}(2020)]%
        {chalkidis_legal-bert_2020}
\bibfield{author}{\bibinfo{person}{Ilias Chalkidis}, \bibinfo{person}{Manos
  Fergadiotis}, \bibinfo{person}{Prodromos Malakasiotis},
  \bibinfo{person}{Nikolaos Aletras}, {and} \bibinfo{person}{Ion
  Androutsopoulos}.} \bibinfo{year}{2020}\natexlab{}.
\newblock \showarticletitle{{LEGAL}-{BERT}: {The} {Muppets} straight out of
  {Law} {School}}.
\newblock \bibinfo{journal}{\emph{arXiv:2010.02559 [cs]}} (\bibinfo{date}{Oct.}
  \bibinfo{year}{2020}).
\newblock
\urldef\tempurl%
\url{http://arxiv.org/abs/2010.02559}
\showURL{%
\tempurl}
\newblock
\shownote{arXiv: 2010.02559}.


\bibitem[Conneau et~al\mbox{.}(2020)]%
        {Conneau.etal2020}
\bibfield{author}{\bibinfo{person}{Alexis Conneau}, \bibinfo{person}{Kartikay
  Khandelwal}, \bibinfo{person}{Naman Goyal}, \bibinfo{person}{Vishrav
  Chaudhary}, \bibinfo{person}{Guillaume Wenzek}, \bibinfo{person}{Francisco
  Guzm{\'a}n}, \bibinfo{person}{Edouard Grave}, \bibinfo{person}{Myle Ott},
  \bibinfo{person}{Luke Zettlemoyer}, {and} \bibinfo{person}{Veselin
  Stoyanov}.} \bibinfo{year}{2020}\natexlab{}.
\newblock \bibinfo{title}{Unsupervised {{Cross-lingual Representation
  Learning}} at {{Scale}}}.
\newblock
\newblock
\urldef\tempurl%
\url{https://doi.org/10.48550/arXiv.1911.02116}
\showDOI{\tempurl}
\showeprint[arxiv]{1911.02116}~[cs]


\bibitem[Devlin et~al\mbox{.}(2019)]%
        {Devlin.etal2019}
\bibfield{author}{\bibinfo{person}{Jacob Devlin}, \bibinfo{person}{Ming-Wei
  Chang}, \bibinfo{person}{Kenton Lee}, {and} \bibinfo{person}{Kristina
  Toutanova}.} \bibinfo{year}{2019}\natexlab{}.
\newblock \showarticletitle{{{BERT}}: {{Pre-training}} of {{Deep Bidirectional
  Transformers}} for {{Language Understanding}}}. In
  \bibinfo{booktitle}{\emph{Proceedings of the 2019 {{Conference}} of the
  {{North American Chapter}} of the {{Association}} for {{Computational
  Linguistics}}: {{Human Language Technologies}}, {{Volume}} 1 ({{Long}} and
  {{Short Papers}})}}. \bibinfo{publisher}{{Association for Computational
  Linguistics}}, \bibinfo{address}{{Minneapolis, Minnesota}},
  \bibinfo{pages}{4171--4186}.
\newblock
\urldef\tempurl%
\url{https://doi.org/10.18653/v1/N19-1423}
\showDOI{\tempurl}


\bibitem[Du et~al\mbox{.}(2019)]%
        {Du}
\bibfield{author}{\bibinfo{person}{Jinhua Du}, \bibinfo{person}{Yan Huang},
  {and} \bibinfo{person}{Karo Moilanen}.} \bibinfo{year}{2019}\natexlab{}.
\newblock \showarticletitle{{{AIG Investments}}.{{AI}} at the {{FinSBD Task}}:
  {{Sentence Boundary Detection}} through {{Sequence Labelling}} and {{BERT
  Fine-tuning}}}. In \bibinfo{booktitle}{\emph{Proceedings of the {{First
  Workshop}} on {{Financial Technology}} and {{Natural Language Processing}}}}.
  \bibinfo{publisher}{{-}}, \bibinfo{address}{{Macao, China}},
  \bibinfo{pages}{81--87}.
\newblock


\bibitem[Gillick(2009)]%
        {Gillick2009}
\bibfield{author}{\bibinfo{person}{Dan Gillick}.}
  \bibinfo{year}{2009}\natexlab{}.
\newblock \showarticletitle{Sentence {{Boundary Detection}} and the {{Problem}}
  with the {{U}}.{{S}}.}. In \bibinfo{booktitle}{\emph{Proceedings of {{Human
  Language Technologies}}: {{The}} 2009 {{Annual Conference}} of the {{North
  American Chapter}} of the {{Association}} for {{Computational Linguistics}},
  {{Companion Volume}}: {{Short Papers}}}}. \bibinfo{publisher}{{Association
  for Computational Linguistics}}, \bibinfo{address}{{Boulder, Colorado}},
  \bibinfo{pages}{241--244}.
\newblock


\bibitem[Gimpel et~al\mbox{.}(2011)]%
        {Gimpel.etal2011}
\bibfield{author}{\bibinfo{person}{Kevin Gimpel}, \bibinfo{person}{Nathan
  Schneider}, \bibinfo{person}{Brendan O'Connor}, \bibinfo{person}{Dipanjan
  Das}, \bibinfo{person}{Daniel Mills}, \bibinfo{person}{Jacob Eisenstein},
  \bibinfo{person}{Michael Heilman}, \bibinfo{person}{Dani Yogatama},
  \bibinfo{person}{Jeffrey Flanigan}, {and} \bibinfo{person}{Noah~A. Smith}.}
  \bibinfo{year}{2011}\natexlab{}.
\newblock \showarticletitle{Part-of-{{Speech Tagging}} for {{Twitter}}:
  {{Annotation}}, {{Features}}, and {{Experiments}}}. In
  \bibinfo{booktitle}{\emph{Proceedings of the 49th {{Annual Meeting}} of the
  {{Association}} for {{Computational Linguistics}}: {{Human Language
  Technologies}}}}. \bibinfo{publisher}{{Association for Computational
  Linguistics}}, \bibinfo{address}{{Portland, Oregon, USA}},
  \bibinfo{pages}{42--47}.
\newblock


\bibitem[Glaser et~al\mbox{.}(2021)]%
        {Glaser}
\bibfield{author}{\bibinfo{person}{Ingo Glaser}, \bibinfo{person}{Sebastian
  Moser}, {and} \bibinfo{person}{Florian Matthes}.}
  \bibinfo{year}{2021}\natexlab{}.
\newblock \showarticletitle{Sentence {{Boundary Detection}} in {{German Legal
  Documents}}:}. In \bibinfo{booktitle}{\emph{Proceedings of the 13th
  {{International Conference}} on {{Agents}} and {{Artificial Intelligence}}}}.
  \bibinfo{publisher}{{SCITEPRESS - Science and Technology Publications}},
  \bibinfo{address}{{Online Streaming, --- Select a Country ---}},
  \bibinfo{pages}{812--821}.
\newblock
\showISBNx{978-989-758-484-8}
\urldef\tempurl%
\url{https://doi.org/10.5220/0010246308120821}
\showDOI{\tempurl}


\bibitem[Goldhahn et~al\mbox{.}(2012)]%
        {Goldhahn.etal2012}
\bibfield{author}{\bibinfo{person}{Dirk Goldhahn}, \bibinfo{person}{Thomas
  Eckart}, {and} \bibinfo{person}{Uwe Quasthoff}.}
  \bibinfo{year}{2012}\natexlab{}.
\newblock \showarticletitle{Building {{Large Monolingual Dictionaries}} at the
  {{Leipzig Corpora Collection}}: {{From}} 100 to 200 {{Languages}}}.
\newblock \bibinfo{journal}{\emph{Proceedings of the Eighth International
  Conference on Language Resources and Evaluation (LREC'12)}}
  \bibinfo{volume}{-}, \bibinfo{number}{-} (\bibinfo{year}{2012}),
  \bibinfo{pages}{--}.
\newblock


\bibitem[Honnibal et~al\mbox{.}(2020)]%
        {spacy}
\bibfield{author}{\bibinfo{person}{Matthew Honnibal}, \bibinfo{person}{Ines
  Montani}, \bibinfo{person}{Sofie Van~Landeghem}, {and}
  \bibinfo{person}{Adriane Boyd}.} \bibinfo{year}{2020}\natexlab{}.
\newblock \showarticletitle{{{spaCy}}: {{Industrial-strength Natural Language
  Processing}} in {{Python}}}.
\newblock \bibinfo{journal}{\emph{-}} \bibinfo{volume}{-}, \bibinfo{number}{-}
  (\bibinfo{year}{2020}), \bibinfo{pages}{--}.
\newblock
\urldef\tempurl%
\url{https://doi.org/10.5281}
\showDOI{\tempurl}


\bibitem[Hua et~al\mbox{.}(2022)]%
        {hua_legalrelectra_2022}
\bibfield{author}{\bibinfo{person}{Wenyue Hua}, \bibinfo{person}{Yuchen Zhang},
  \bibinfo{person}{Zhe Chen}, \bibinfo{person}{Josie Li}, {and}
  \bibinfo{person}{Melanie Weber}.} \bibinfo{year}{2022}\natexlab{}.
\newblock \bibinfo{title}{{LegalRelectra}: {Mixed}-domain {Language} {Modeling}
  for {Long}-range {Legal} {Text} {Comprehension}}.
\newblock
\newblock
\urldef\tempurl%
\url{https://doi.org/10.48550/arXiv.2212.08204}
\showDOI{\tempurl}
\newblock
\shownote{arXiv:2212.08204 [cs]}.


\bibitem[Katz et~al\mbox{.}(2023)]%
        {Katz.etal2023}
\bibfield{author}{\bibinfo{person}{Daniel~Martin Katz}, \bibinfo{person}{Dirk
  Hartung}, \bibinfo{person}{Lauritz Gerlach}, \bibinfo{person}{Abhik Jana},
  {and} \bibinfo{person}{Michael~James Bommarito}.}
  \bibinfo{year}{2023}\natexlab{}.
\newblock \bibinfo{title}{Natural {{Language Processing}} in the {{Legal
  Domain}}}.
\newblock
\newblock


\bibitem[Kiss and Strunk(2006)]%
        {Kiss}
\bibfield{author}{\bibinfo{person}{Tibor Kiss} {and} \bibinfo{person}{Jan
  Strunk}.} \bibinfo{year}{2006}\natexlab{}.
\newblock \showarticletitle{Unsupervised {{Multilingual Sentence Boundary
  Detection}}}.
\newblock \bibinfo{journal}{\emph{Computational Linguistics}}
  \bibinfo{volume}{32}, \bibinfo{number}{4} (\bibinfo{date}{Dec.}
  \bibinfo{year}{2006}), \bibinfo{pages}{485--525}.
\newblock
\showISSN{0891-2017, 1530-9312}
\urldef\tempurl%
\url{https://doi.org/10.1162/coli.2006.32.4.485}
\showDOI{\tempurl}


\bibitem[Koehn(2005)]%
        {koehnEuroparlParallelCorpus2005}
\bibfield{author}{\bibinfo{person}{Philipp Koehn}.}
  \bibinfo{year}{2005}\natexlab{}.
\newblock \showarticletitle{Europarl: {{A Parallel Corpus}} for {{Statistical
  Machine Translation}}}. In \bibinfo{booktitle}{\emph{Proceedings of {{Machine
  Translation Summit X}}: {{Papers}}}}, Vol.~\bibinfo{volume}{-}.
  \bibinfo{publisher}{{-}}, \bibinfo{address}{{Phuket, Thailand}},
  \bibinfo{pages}{79--86}.
\newblock


\bibitem[Lin et~al\mbox{.}(2020)]%
        {Lin.etal2020}
\bibfield{author}{\bibinfo{person}{Chun-Wei Lin}, \bibinfo{person}{Yinan Shao},
  \bibinfo{person}{Ji Zhang}, {and} \bibinfo{person}{Unil Yun}.}
  \bibinfo{year}{2020}\natexlab{}.
\newblock \showarticletitle{Enhanced {{Sequence Labeling Based}} on {{Latent
  Variable Conditional Random Fields}}}.
\newblock \bibinfo{journal}{\emph{Neurocomputing}} \bibinfo{volume}{403},
  \bibinfo{number}{-} (\bibinfo{date}{May} \bibinfo{year}{2020}),
  \bibinfo{pages}{--}.
\newblock
\urldef\tempurl%
\url{https://doi.org/10.1016/j.neucom.2020.04.102}
\showDOI{\tempurl}


\bibitem[Manning et~al\mbox{.}(2014)]%
        {CorenlpManning}
\bibfield{author}{\bibinfo{person}{Christopher Manning}, \bibinfo{person}{Mihai
  Surdeanu}, \bibinfo{person}{John Bauer}, \bibinfo{person}{Jenny Finkel},
  \bibinfo{person}{Steven Bethard}, {and} \bibinfo{person}{David McClosky}.}
  \bibinfo{year}{2014}\natexlab{}.
\newblock \showarticletitle{The {{Stanford CoreNLP Natural Language Processing
  Toolkit}}}. In \bibinfo{booktitle}{\emph{Proceedings of 52nd {{Annual
  Meeting}} of the {{Association}} for {{Computational Linguistics}}: {{System
  Demonstrations}}}}. \bibinfo{publisher}{{Association for Computational
  Linguistics}}, \bibinfo{address}{{Baltimore, Maryland}},
  \bibinfo{pages}{55--60}.
\newblock
\urldef\tempurl%
\url{https://doi.org/10.3115/v1/P14-5010}
\showDOI{\tempurl}


\bibitem[Mathew and Guggilla(2019)]%
        {Mathew.Guggilla2019}
\bibfield{author}{\bibinfo{person}{Ditty Mathew} {and}
  \bibinfo{person}{Chinnappa Guggilla}.} \bibinfo{year}{2019}\natexlab{}.
\newblock \showarticletitle{{{AI}}\_{{Blues}} at {{FinSBD Shared Task}}:
  {{CRF-based Sentence Boundary Detection}} in {{PDF Noisy Text}} in the
  {{Financial Domain}}}. In \bibinfo{booktitle}{\emph{Proceedings of the
  {{First Workshop}} on {{Financial Technology}} and {{Natural Language
  Processing}}}}. \bibinfo{publisher}{{-}}, \bibinfo{address}{{Macao, China}},
  \bibinfo{pages}{130--136}.
\newblock


\bibitem[{Newman-Griffis} et~al\mbox{.}(2016)]%
        {Newman-Griffis.etal2016}
\bibfield{author}{\bibinfo{person}{Denis {Newman-Griffis}},
  \bibinfo{person}{Chaitanya Shivade}, \bibinfo{person}{Eric {Fosler-Lussier}},
  {and} \bibinfo{person}{Albert Lai}.} \bibinfo{year}{2016}\natexlab{}.
\newblock \showarticletitle{A {{Quantitative}} and {{Qualitative Evaluation}}
  of {{Sentence Boundary Detection}} for the {{Clinical Domain}}}.
\newblock \bibinfo{journal}{\emph{AMIA Joint Summits on Translational Science
  proceedings. AMIA Summit on Translational Science}}  \bibinfo{volume}{2016}
  (\bibinfo{date}{July} \bibinfo{year}{2016}), \bibinfo{pages}{88--97}.
\newblock


\bibitem[Niklaus et~al\mbox{.}(2021)]%
        {niklaus_swiss-judgment-prediction_2021}
\bibfield{author}{\bibinfo{person}{Joel Niklaus}, \bibinfo{person}{Ilias
  Chalkidis}, {and} \bibinfo{person}{Matthias Stürmer}.}
  \bibinfo{year}{2021}\natexlab{}.
\newblock \showarticletitle{Swiss-{Judgment}-{Prediction}: {A} {Multilingual}
  {Legal} {Judgment} {Prediction} {Benchmark}}. In
  \bibinfo{booktitle}{\emph{Proceedings of the {Natural} {Legal} {Language}
  {Processing} {Workshop} 2021}}. \bibinfo{publisher}{Association for
  Computational Linguistics}, \bibinfo{address}{Punta Cana, Dominican
  Republic}, \bibinfo{pages}{19--35}.
\newblock
\urldef\tempurl%
\url{https://aclanthology.org/2021.nllp-1.3}
\showURL{%
\tempurl}


\bibitem[Niklaus and Giofré(2022)]%
        {niklaus_budgetlongformer_2022}
\bibfield{author}{\bibinfo{person}{Joel Niklaus} {and} \bibinfo{person}{Daniele
  Giofré}.} \bibinfo{year}{2022}\natexlab{}.
\newblock \bibinfo{title}{{BudgetLongformer}: {Can} we {Cheaply} {Pretrain} a
  {SotA} {Legal} {Language} {Model} {From} {Scratch}?}
\newblock
\newblock
\urldef\tempurl%
\url{https://doi.org/10.48550/arXiv.2211.17135}
\showDOI{\tempurl}
\newblock
\shownote{arXiv:2211.17135 [cs]}.


\bibitem[Niklaus et~al\mbox{.}(2023)]%
        {niklaus_lextreme_2023}
\bibfield{author}{\bibinfo{person}{Joel Niklaus}, \bibinfo{person}{Veton
  Matoshi}, \bibinfo{person}{Pooja Rani}, \bibinfo{person}{Andrea Galassi},
  \bibinfo{person}{Matthias Stürmer}, {and} \bibinfo{person}{Ilias
  Chalkidis}.} \bibinfo{year}{2023}\natexlab{}.
\newblock \bibinfo{title}{{LEXTREME}: {A} {Multi}-{Lingual} and {Multi}-{Task}
  {Benchmark} for the {Legal} {Domain}}.
\newblock
\newblock
\urldef\tempurl%
\url{https://doi.org/10.48550/arXiv.2301.13126}
\showDOI{\tempurl}
\newblock
\shownote{arXiv:2301.13126 [cs]}.


\bibitem[Niklaus et~al\mbox{.}(2022)]%
        {niklaus_empirical_2022}
\bibfield{author}{\bibinfo{person}{Joel Niklaus}, \bibinfo{person}{Matthias
  Stürmer}, {and} \bibinfo{person}{Ilias Chalkidis}.}
  \bibinfo{year}{2022}\natexlab{}.
\newblock \showarticletitle{An {Empirical} {Study} on {Cross}-{X} {Transfer}
  for {Legal} {Judgment} {Prediction}}. In
  \bibinfo{booktitle}{\emph{Proceedings of the 2nd {Conference} of the
  {Asia}-{Pacific} {Chapter} of the {Association} for {Computational}
  {Linguistics} and the 12th {International} {Joint} {Conference} on {Natural}
  {Language} {Processing} ({Volume} 1: {Long} {Papers})}}.
  \bibinfo{publisher}{Association for Computational Linguistics},
  \bibinfo{address}{Online only}, \bibinfo{pages}{32--46}.
\newblock
\urldef\tempurl%
\url{https://aclanthology.org/2022.aacl-main.3}
\showURL{%
\tempurl}


\bibitem[Qi et~al\mbox{.}(2020)]%
        {Qi.etal2020a}
\bibfield{author}{\bibinfo{person}{Peng Qi}, \bibinfo{person}{Yuhao Zhang},
  \bibinfo{person}{Yuhui Zhang}, \bibinfo{person}{Jason Bolton}, {and}
  \bibinfo{person}{Christopher~D. Manning}.} \bibinfo{year}{2020}\natexlab{}.
\newblock \showarticletitle{Stanza: {{A Python Natural Language Processing
  Toolkit}} for {{Many Human Languages}}}. In
  \bibinfo{booktitle}{\emph{Proceedings of the 58th {{Annual Meeting}} of the
  {{Association}} for {{Computational Linguistics}}: {{System
  Demonstrations}}}}. \bibinfo{publisher}{{Association for Computational
  Linguistics}}, \bibinfo{address}{{Online}}, \bibinfo{pages}{101--108}.
\newblock
\urldef\tempurl%
\url{https://doi.org/10.18653/v1/2020.acl-demos.14}
\showDOI{\tempurl}


\bibitem[Read et~al\mbox{.}(2012)]%
        {Read}
\bibfield{author}{\bibinfo{person}{Jonathon Read}, \bibinfo{person}{Rebecca
  Dridan}, \bibinfo{person}{Stephan Oepen}, {and} \bibinfo{person}{Lars~Jrgen
  Solberg}.} \bibinfo{year}{2012}\natexlab{}.
\newblock \showarticletitle{Sentence {{Boundary Detection}}: {{A Long Solved
  Problem}}?}. In \bibinfo{booktitle}{\emph{Proceedings of {{COLING}} 2012:
  {{Posters}}}}. \bibinfo{publisher}{{-}}, \bibinfo{address}{{Mumbai, India}},
  \bibinfo{pages}{985--994}.
\newblock


\bibitem[Sanchez(2019)]%
        {Sanchez}
\bibfield{author}{\bibinfo{person}{George Sanchez}.}
  \bibinfo{year}{2019}\natexlab{}.
\newblock \showarticletitle{Sentence {{Boundary Detection}} in {{Legal Text}}}.
  In \bibinfo{booktitle}{\emph{Proceedings of the {{Natural Legal Language
  Processing Workshop}} 2019}}. \bibinfo{publisher}{{Association for
  Computational Linguistics}}, \bibinfo{address}{{Minneapolis, Minnesota}},
  \bibinfo{pages}{31--38}.
\newblock
\urldef\tempurl%
\url{https://doi.org/10.18653/v1/W19-2204}
\showDOI{\tempurl}


\bibitem[Sanh et~al\mbox{.}(2019)]%
        {Distil}
\bibfield{author}{\bibinfo{person}{Victor Sanh}, \bibinfo{person}{Lysandre
  Debut}, \bibinfo{person}{Julien Chaumond}, {and} \bibinfo{person}{Thomas
  Wolf}.} \bibinfo{year}{2019}\natexlab{}.
\newblock \showarticletitle{{{DistilBERT}}, a Distilled Version of {{BERT}}:
  Smaller, Faster, Cheaper and Lighter}.
\newblock \bibinfo{journal}{\emph{ArXiv}} \bibinfo{volume}{abs/1910.01108},
  \bibinfo{number}{-} (\bibinfo{year}{2019}), \bibinfo{pages}{--}.
\newblock


\bibitem[Savelka et~al\mbox{.}(2017)]%
        {Savelka}
\bibfield{author}{\bibinfo{person}{Jaromir Savelka}, \bibinfo{person}{Vern~R
  Walker}, \bibinfo{person}{Matthias Grabmair}, {and} \bibinfo{person}{Kevin~D
  Ashley}.} \bibinfo{year}{2017}\natexlab{}.
\newblock \showarticletitle{Sentence {{Boundary Detection}} in {{Adjudicatory
  Decisions}} in the {{United States}}}.
\newblock \bibinfo{journal}{\emph{TAL}} \bibinfo{volume}{58},
  \bibinfo{number}{21} (\bibinfo{year}{2017}), \bibinfo{pages}{--}.
\newblock


\bibitem[Savelka et~al\mbox{.}(2021)]%
        {Savelka.etal2021}
\bibfield{author}{\bibinfo{person}{Jaromir Savelka}, \bibinfo{person}{Hannes
  Westermann}, \bibinfo{person}{Karim Benyekhlef},
  \bibinfo{person}{Charlotte~S. Alexander}, \bibinfo{person}{Jayla~C. Grant},
  \bibinfo{person}{David~Restrepo Amariles}, \bibinfo{person}{Rajaa~El
  Hamdani}, \bibinfo{person}{S{\'e}bastien Mee{\`u}s}, \bibinfo{person}{Aurore
  Troussel}, \bibinfo{person}{Micha{\l} Araszkiewicz},
  \bibinfo{person}{Kevin~D. Ashley}, \bibinfo{person}{Alexandra Ashley},
  \bibinfo{person}{Karl Branting}, \bibinfo{person}{Mattia Falduti},
  \bibinfo{person}{Matthias Grabmair}, \bibinfo{person}{Jakub Hara{\v s}ta},
  \bibinfo{person}{Tereza Novotn{\'a}}, \bibinfo{person}{Elizabeth Tippett},
  {and} \bibinfo{person}{Shiwanni Johnson}.} \bibinfo{year}{2021}\natexlab{}.
\newblock \showarticletitle{Lex {{Rosetta}}: Transfer of Predictive Models
  across Languages, Jurisdictions, and Legal Domains}. In
  \bibinfo{booktitle}{\emph{Proceedings of the {{Eighteenth International
  Conference}} on {{Artificial Intelligence}} and {{Law}}}}.
  \bibinfo{publisher}{{ACM}}, \bibinfo{address}{{S\~ao Paulo Brazil}},
  \bibinfo{pages}{129--138}.
\newblock
\showISBNx{978-1-4503-8526-8}
\urldef\tempurl%
\url{https://doi.org/10.1145/3462757.3466149}
\showDOI{\tempurl}


\bibitem[Schweter and Ahmed(2019)]%
        {SchweterAhmed}
\bibfield{author}{\bibinfo{person}{Stefan Schweter} {and}
  \bibinfo{person}{Sajawel Ahmed}.} \bibinfo{year}{2019}\natexlab{}.
\newblock \showarticletitle{Deep-{{EOS}}: {{General-Purpose Neural Networks}}
  for {{Sentence Boundary Detection}}}. In
  \bibinfo{booktitle}{\emph{Proceedings of the 15th {{Conference}} on {{Natural
  Language Processing}} ({{KONVENS}})}}, Vol.~\bibinfo{volume}{-}.
  \bibinfo{publisher}{{-}}, \bibinfo{address}{{-}}, \bibinfo{pages}{5}.
\newblock


\bibitem[Song et~al\mbox{.}(2020)]%
        {Song.etal2020}
\bibfield{author}{\bibinfo{person}{Xinying Song}, \bibinfo{person}{Alex
  Salcianu}, \bibinfo{person}{Yang Song}, \bibinfo{person}{Dave Dopson}, {and}
  \bibinfo{person}{Denny Zhou}.} \bibinfo{year}{2020}\natexlab{}.
\newblock \showarticletitle{Fast {{WordPiece Tokenization}}}.
\newblock \bibinfo{journal}{\emph{-}} \bibinfo{volume}{-}, \bibinfo{number}{-}
  (\bibinfo{year}{2020}), \bibinfo{pages}{--}.
\newblock
\urldef\tempurl%
\url{https://doi.org/10.48550/ARXIV.2012.15524}
\showDOI{\tempurl}


\bibitem[Tiedemann(2012)]%
        {Tiedemann2012}
\bibfield{author}{\bibinfo{person}{Jorg Tiedemann}.}
  \bibinfo{year}{2012}\natexlab{}.
\newblock \showarticletitle{Parallel {{Data}}, {{Tools}} and {{Interfaces}} in
  {{OPUS}}}.
\newblock \bibinfo{journal}{\emph{In Proceedings of the 8th International
  Conference on Language Resources and Evaluation (LREC 2012)}}
  \bibinfo{volume}{-}, \bibinfo{number}{-} (\bibinfo{year}{2012}),
  \bibinfo{pages}{--}.
\newblock


\bibitem[Vaswani et~al\mbox{.}(2017)]%
        {Vaswani.etal2017}
\bibfield{author}{\bibinfo{person}{Ashish Vaswani}, \bibinfo{person}{Noam
  Shazeer}, \bibinfo{person}{Niki Parmar}, \bibinfo{person}{Jakob Uszkoreit},
  \bibinfo{person}{Llion Jones}, \bibinfo{person}{Aidan~N. Gomez},
  \bibinfo{person}{Lukasz Kaiser}, {and} \bibinfo{person}{Illia Polosukhin}.}
  \bibinfo{year}{2017}\natexlab{}.
\newblock \bibinfo{title}{Attention {{Is All You Need}}}.
\newblock
\newblock
\urldef\tempurl%
\url{https://doi.org/10.48550/arXiv.1706.03762}
\showDOI{\tempurl}
\showeprint[arxiv]{1706.03762}~[cs]


\end{thebibliography}

%\appendix
%\input{Content/Appendix.tex}

\end{document}